\documentclass[preprint,1p,10pt]{elsarticle}

\usepackage{lipsum}
\usepackage{graphicx}
\usepackage{amsmath, amssymb}
\usepackage{amsfonts}
\usepackage{graphicx,adjustbox}
\usepackage{caption}
\usepackage{subcaption}
\usepackage[colorlinks,bookmarksopen,bookmarksnumbered,citecolor=red,urlcolor=red]{hyperref}
\usepackage[per-mode=symbol]{siunitx}
\usepackage[english]{babel}
\usepackage{tikz}
\usetikzlibrary{quantikz}
\usepackage[T1]{fontenc}
\usepackage{float}
\usepackage[utf8]{inputenc}
\usepackage{lmodern}
\usepackage{wrapfig}
\usepackage{mathtools}
\usepackage{enumitem}
\usepackage{braket}
\usepackage{booktabs}
\usepackage[noprefix]{nomencl}
\makenomenclature
\usepackage{lineno}
\usepackage{color}
\usepackage{threeparttable}
\usepackage{multirow}

\usepackage{ulem}

\biboptions{sort&compress}

\journal{XXXXXXXX}

\begin{document}

\begin{frontmatter}

\title{Quantum Fourier Transform Based Kernel for Solar Irrandiance Forecasting}

\author[EPFL]{Nawfel Mechiche-Alami\texorpdfstring{\corref{cor1}}{}}
\ead{nawfel.mechiche-alami@epfl.ch}
\cortext[cor1]{Corresponding author}
\author[UCH]{Eduardo Rodríguez}
\ead{edrodriguez@ug.uchile.cl}
\author[PUC]{José M. Cardemil}
\author[UCLA2,UCLA]{Enrique L\'{o}pez Droguett}

\address[EPFL]{Ecole Polytechnique Fédérale de Lausanne (EPFL), CH-1015 Lausanne, Switzerland}
\address[UCH]{Mechanical Engineering Department, Universidad de Chile, Santiago 8370456, Chile}
\address[PUC]{Departamento de Ingeniería Mecánica y Metalúrgica, Escuela de Ingeniería, Pontificia Universidad
Católica de Chile, Vicuña Mackenna 4860, Santiago, Chile}
\address[UCLA2]{Garrick Institute for the Risk Sciences, University of California, Los Angeles, CA 90095, USA}
\address[UCLA]{Dept. of Civil and Environmental Engineering, Univ. of California, Los Angeles, CA 90095, USA}

\begin{abstract}

This study proposes a Quantum Fourier Transform (QFT)–enhanced quantum kernel for short-term time-series forecasting. Each signal is windowed, amplitude-encoded, transformed by a QFT, then passed through a protective rotation layer to avoid the QFT/QFT$^\dagger$ cancellation; the resulting kernel is used in kernel ridge regression (KRR). Exogenous predictors are incorporated by convexly fusing feature-specific kernels. On multi-station solar irradiance data across Köppen climate classes, the proposed kernel consistently improves median $R^2$ and $n$RMSE over reference classical RBF and polynomial kernels, while also reducing bias ($n$MBE); complementary MAE/ERMAX analyses indicate tighter average errors with remaining headroom under sharp transients. For both quantum and classical models, the only tuned quantities are the feature-mixing weights and the KRR ridge $\alpha$; classical hyperparameters $(\gamma,r,d)$ are fixed, with the same validation set size for all models. Experiments are conducted on a noiseless simulator (5 qubits; window length $L{=}32$). Limitations and ablations are discussed, and paths toward NISQ execution are outlined.

\end{abstract}

\begin{keyword}
Quantum kernels \sep Solar irradiance \sep Forecasting.
\end{keyword}

\end{frontmatter}

\section{Introduction}


Quantum Machine Learning (QML) is an emerging discipline that combines the principles of quantum physics with traditional machine learning (ML) to exploit the distinctive characteristics of quantum systems, including superposition and entanglement phenomena \cite{biamonte2017quantum}. Conventional machine learning algorithms use the bits of classical computers, whereas quantum machine learning employs quantum bits (qubits) from quantum computers, facilitating activities such as data parallelization in superposition states. This distinction facilitates the expeditious execution of certain tasks \cite{harrow2009quantum}, such as classification and dimensionality reduction, where QML has demonstrated significant acceleration \cite{rebentrost2014quantum}.

QML applications have extended to time-series data, leveraging quantum phenomena to model complex temporal dependencies. The goal is to enhance the results of traditional tasks by performing computations on qubits, which can process data more efficiently than classical bits \cite{Hirth_Droguett_2025, Takaki_Mitarai_Negoro_Fujii_Kitagawa_2021}. For example, \citeauthor{Thakkar_2024} \cite{Thakkar_2024} demonstrated that quantum machine-learning methods could enhance financial forecasting by improving both churn prediction and credit-risk assessment. Likewise, \citeauthor{Kea_Kim_Huot_Kim_Han_2024} \cite{Kea_Kim_Huot_Kim_Han_2024} developed a hybrid quantum-classical Long Short-Term Memory (QLSTM) to improve stock-price forecasting by leveraging quantum data encoding and high-dimensional quantum representations. In the energy domain, \citeauthor{Hangun_Akpinar_Altun_Eyecioglu_2025} \cite{Hangun_Akpinar_Altun_Eyecioglu_2025} evaluated how Quantum Neural Network architectures performed in wind-power forecasting by systematically varying feature maps and ansatz entanglement strategies. \citeauthor{Ceschini_Rosato_Panella_2022} \cite{Ceschini_Rosato_Panella_2022} applied quantum-hybrid Recurrent Neural Networks (QRNNs) to enhance time-series forecasting for photovoltaic power to capture seasonality and weather conditions. Its goal was to show that this hybrid model improved prediction accuracy for photovoltaic power time series compared to classical LSTM and Random Forest baselines. 

Quantum kernels are functions designed to map classical data into a high-dimensional quantum Hilbert space, enabling measurement-based computation of similarities between data points \cite{havlicek2019supervised}. Classical kernel methods rely on mathematically defined feature mappings that may become inefficient as dimensionality increases. Quantum kernels, however, encode classical information into quantum states and compute inner products through quantum measurements, potentially bypassing the need for explicit feature-map computation \cite{Gil_Fuster_2024}. Several researchers have proposed quantum kernel classifiers in which a quantum circuit encodes classical time-series data into quantum states, enabling a kernel function to be estimated via inner products of these states. In particular, hybrid quantum-classical models have been developed where temporal dependencies are implicitly captured by encoding entire sequences into quantum feature maps, and classification is performed by evaluating similarities between encoded sequences using quantum kernels \cite{baker_parallel_2024}. In the context of time series forecasting, researchers have developed quantum-kernel based regression models in which classical temporal sequences are embedded into quantum feature states via a parameterized quantum circuit, and a kernel function, evaluated as the overlap of those quantum states, captures the temporal dependencies among sequence data. This hybrid quantum-classical formulation aims to improve prediction of future values by exploiting the representation power of the quantum feature map for temporal correlations and patterns \cite{Aaraba_2024}.

The Quantum Fourier Transform (QFT) is the quantum analogue of the classical Discrete Fourier Transform (DFT), operating on quantum states rather than classical signals. It serves as a core subroutine in various quantum algorithms, most notably in Shor’s factoring algorithm \cite{shor1994algorithms}. By exploiting quantum parallelism, the QFT can be computed with $\mathcal{O}(n^2)$ gate operations for an $n$-qubit state, which is exponentially faster than classical Fast Fourier Transform (FFT) methods when accounting for the number of input states processed in superposition \cite{nielsen2010quantum}. Although its asymptotic circuit complexity can be favorable, it does not necessarily imply practical speedups for classical datasets, where state preparation and measurement overhead dominate. In this study, the QFT is used as a representation to expose periodic structure; accuracy is evaluated rather than runtime benefits.

The success of QFT has motivated exploration of QFT-based techniques in other domains, including the analysis of periodic signals within time-series data \cite{shor1994algorithms}. Researchers have explored the use of the Quantum Fourier Transform (QFT) to decompose time-series signals into frequency components by exploiting quantum parallelism and superposition. In these approaches, temporal sequences are amplitude-encoded into quantum states, and a QFT circuit maps the time-domain representation into a compact spectral domain where dominant frequency patterns become more explicit. Such QFT-based techniques aim to enhance the extraction of multi-scale periodic structures in complex time series, providing frequency-domain features that can support downstream forecasting models \cite{Tang_Cai_Zhang_Gao_Yu_2025}. However, the practical implementation of QFT-based time-series analysis faces several challenges, including the efficient loading of large classical datasets into quantum states and the mitigation of noise on current quantum hardware \cite{Marin_Sanchez_2023}.

Time series data are inherently dynamic because of their temporal fluctuations. They exhibit intrinsic characteristics that make traditional methods for static data insufficient to uncover hidden dependencies \cite{barandas2020}. Such data frequently display trends, seasonality, and cyclical patterns, resulting in variations in statistical properties such as mean and variance over time. Conventional approaches typically assume stationarity, making them inadequate for capturing the dynamic patterns intrinsic to time series data. Consequently, traditional techniques often struggle to extract meaningful features from time series data, which differ substantially from datasets composed of independent and identically distributed observations. Specialized approaches for time-series analysis are therefore essential \cite{kim2024}.

Time series are particularly relevant in the field of solar energy forecasting. Accurate forecasting is critical for the effective operation and maintenance of solar installations \cite{bouadjila2024}. Predicting solar radiation mitigates the uncertainty of energy generation and supports informed decision-making \cite{kaur2016}. Meteorological variables fluctuate continuously, making prediction in this field especially challenging. Numerous studies have investigated solar radiation forecasting, reflecting the growing academic interest in developing more precise models \cite{lai2021, bansal2023, wen2023, despotovic2024, sanchez2024, li2025}. Forecasting horizons are generally categorized as short-term, medium-term, and long-term. Short-term forecasting is the most challenging, as it often requires predictions at minute-level intervals.

Given the potential of quantum kernels and the scarcity of existing quantum kernel approaches for time-series forecasting, the present study aims to design a novel quantum kernel explicitly tailored for short-term time-series forecasting. By integrating the Quantum Fourier Transform (QFT) into the quantum kernel, this research leverages its capacity to analyze frequency components within time-series data, potentially enhancing forecasting capability. The study extends the application of quantum kernels to dynamic data and explores strategies to improve forecasting accuracy. In doing so, it addresses a gap in the literature and lays the foundation for further advances in quantum-enhanced forecasting.

The paper is organized as follows: initially, it is provided a review on the QML used in the literature to predict solar radiation is provided. Then, the proposed approach, followed by the methodology for QFT is established. The results of the predictions for several meteorological stations are shown, followed by a comparison using a traditional machine learning model. A discussion of the outcomes is ultimately included. 

\section{QML for Estimating Solar Radiation}

To the best of the authors' knowledge, the 2016 study by \citeauthor{senekane2016} \cite{senekane2016} was the first to apply QML in the field of solar radiation. To forecast solar irradiance, \citeauthor{senekane2016} employed quantum machine learning techniques on meteorological data obtained from the Digital Technology Group (DTG) Weather Station at the University of Cambridge \cite{DTG_WeatherStation}. The dataset included three input features: temperature, humidity, and wind speed, and consisted of 49 instances recorded at 30-minute intervals. The quantum model used a half-hourly timestep, with global irradiance as the target variable for prediction. A critical preprocessing step involved converting classical information into quantum states to enable quantum computations. Matrix inversion techniques were then applied to optimize the parameters of the quantum support vector hyperplane using a quantum algorithm.

The model proposed by \citeauthor{senekane2016} employs a Quantum Support Vector Machine (QSVM) approach that reduces the optimization problem to a system of linear equations, leveraging a quantum linear-systems solver with matrix inversion that is known to provide exponential speedup at the subroutine level. During post-processing, prediction errors such as mean squared error (MSE), root mean squared error (RMSE), mean absolute error (MAE), and the coefficient of determination ($R^2$) were calculated for different training sizes. Using 70\% of the dataset for training produced the best performance, with an RMSE of 1.626 and an $R^2$ of 0.852. These results underline the robustness and illustrate the utility of the proposed QSVM for solar irradiation prediction.

In 2023, \citeauthor{sushmit2023} \cite{sushmit2023} used hourly meteorological data from NASA’s POWER project for the Rangamati region in Bangladesh, covering the period from January 2010 to December 2019. The dataset contained 16 metrics, including all-sky insolation fraction, solar zenith angle, surface temperature, relative humidity, and wind speed. Seven features were selected for model training by first using Pearson correlation to identify the top 10 and then refining to seven based on prior literature, with solar irradiance as the target variable. Validation was performed using 93,171 data points from eight global locations spanning 2020 to 2023, obtained from an independent dataset. As a preprocessing step, all features were scaled to the range [$-1,1$] to improve training performance.

To establish a baseline for comparison, \citeauthor{sushmit2023} implemented a bidirectional long short-term memory (BiLSTM) model to evaluate the performance of the remaining models. A hybrid classical–quantum approach incorporating Parameterized Quantum Circuits (PQC) was then employed to develop predictive architectures, such as feedforward neural networks (FFNs) integrated with quantum layers at various depths. The models FFN4L1Q, FFN5L2Q, and FFN8L1Q used quantum gates ($R_X$, $R_Y$, $R_Z$, and CNOT) for angle encoding and entanglement, each with one or two quantum layers. The PQC operated entirely within a five-layer quantum framework. All models were trained using the ADAM optimizer for 200 epochs. Evaluation metrics including MAE, MSE, RMSE, and $R^2$ were computed to assess predictive performance. The BiLSTM model achieved the best overall accuracy, with an RMSE of 11.052 and an $R^2$ of 0.998. The FFN8L1Q model performed comparably, reaching an RMSE of 12.166 and an $R^2$ of 0.998, while the FFN5L2Q model, despite using far fewer parameters (339), showed a higher error with an RMSE of 24.587 and an $R^2$ of 0.991. These results illustrate that although the classical BiLSTM remains superior in accuracy, with 49,666 parameters, the hybrid FFN8L1Q achieved nearly equivalent performance using only 5,551 parameters, demonstrating the potential of quantum–classical architectures to deliver competitive results with significantly reduced model complexity.

The study conducted by \citeauthor{yu2023} \cite{yu2023} used hourly meteorological data from the U.S. National Solar Radiation Database (NSRDB) for five cities in China, covering the period from January 2016 to December 2019. The dataset included eight variables: global horizontal irradiance (GHI, $W/m^2$), temperature, pressure, relative humidity, wind speed, cloud type, dew point, and solar zenith angle. The target variable was GHI, which was used for both model training and validation. Inputs were Min–Max normalized to [0,1], and predicted values were denormalized using the same Min–Max method. The dataset was divided into training set (2016-2018) and testing set (2019) based on chronological order, ensuring that the test data followed the training period. 

\citeauthor{yu2023} developed a Quantum Long Short-Term Memory (QLSTM) network that integrates a variational quantum circuit (VQC) with a classical long short-term memory (LSTM) architecture. Each QLSTM cell contains four VQCs: one for the forget gate (one-dimensional information flow), one for the input gate (multi-dimensional information flow), one for the cell state (temporal decision-making), and one for the output gate (time-dependent regulation of memory), enabling the model to capture complex temporal features. The architecture combines quantum feature embedding, variational layers with entanglement mechanisms, and classical LSTM operations. During post-processing, the predicted values were denormalized using the Min–Max method. The model’s performance was evaluated using RMSE, MAE, and $R^2$. The QLSTM achieved an average annual RMSE of 61.756 $W/m^2$, an MAE of 24.257 $W/m^2$, and an $R^2$ of 0.946. Compared with baseline models such as SARIMA, CNN, RNN, GRU, and conventional LSTM, the QLSTM demonstrated substantial improvements in predictive accuracy across all evaluation metrics.

The study presented by \citeauthor{oliveira2024} \cite{oliveira2024} used the Folsom dataset \cite{pedro_2019}, which contains minute-by-minute ground-based measurements of global horizontal irradiance (GHI) and direct normal irradiance (DNI) collected in California, USA. The dataset spans the period from 2014 to 2016 and is complemented by sky images and satellite observations. The forecasting task targets both GHI and DNI with prediction horizons ranging from 5 minutes to 3 hours. To improve prediction quality, the data underwent preprocessing steps that included normalization using clear-sky index modeling and recursive feature elimination (RFE) for optimal feature selection. Eighteen input variables were considered, including backward and lagged averages, clear-sky variability, and sky image attributes such as the mean, standard deviation, and entropy for the red, green, and blue channels.

The model proposed by \citeauthor{oliveira2024} employed a Quantum Neural Network (QNN) framework developed with Qiskit, integrating Pauli-Y angle encoding as the feature map and a Two-Local ansatz with controlled-NOT (CNOT) gates for linear entanglement. The L-BFGS-B optimizer was selected for its superior performance in minimizing the cost function. The results indicated that while classical models such as XGBoost outperformed the QNN for shorter horizons, the QNN exhibited greater robustness for longer prediction intervals. Specifically, for a 3-hour horizon, the QNN achieved for GHI an RMSE of 77.55 $W/m^2$, $R^2$ of 94.63\%, and Forecast Skill of 80.92\%; and for DNI an RMSE of 168.81 $W/m^2$, $R^2$ of 80.64\%, and Forecast Skill of 66.22\%, ranking as the second-best model among those evaluated. The study underscores the potential of quantum machine learning for extended forecasting tasks, demonstrating competitive performance relative to well-established classical approaches.

Although kernel-based approaches can capture complex nonlinear relationships in data, none of the reviewed studies incorporated quantum kernels into their quantum models. Furthermore, the Quantum Fourier Transform (QFT), which has been successfully applied in other fields, has not yet been used for forecasting solar radiation. The QFT is well known for its ability to transform data into the frequency domain, enabling the extraction of periodic patterns and harmonic components from solar irradiance time series. This capability could enhance the predictive performance of quantum models by improving their ability to represent seasonal and diurnal cycles, which are critical in solar energy applications. Consequently, integrating QFT and quantum kernels into quantum models provides a promising direction for increasing the accuracy of solar irradiance forecasting.

\section{Proposed Approach}


The proposed method entails creating an innovative quantum kernel specifically designed for short-term time series forecasting via the use of the QFT. The QFT excels in analyzing frequency components in data, rendering it particularly suitable for time series that display periodic or quasi-periodic characteristics. By integrating the QFT into the quantum kernel, the approach seeks to more effectively capture the fundamental frequency patterns compared to conventional methods. This integration allows a more sophisticated depiction of the time series inside the quantum feature space, possibly resulting in enhanced accuracy and reliability of predictions.

The use of the QFT-enhanced quantum kernel has several advantages. It improves the model's capacity to identify and anticipate intricate temporal patterns present in time series data, which is especially beneficial in applications such as solar energy forecasting, where short-term forecasts are essential. Enhanced forecasting precision may result in more effective energy management and superior decision-making processes. This methodology broadens the use of quantum kernels beyond static data categorization. 

\subsection{Solar data}

The data employed for the forecasting model were sourced from meteorological stations that are part of the Baseline Surface Radiation Network (BSRN) \cite{wrmc-bsrn_world_2018, driemel_baseline_2018}. These BSRN stations are recognized as first-quality stations, providing highly reliable data stored at a fine temporal resolution of one minute. In this study, a total of 30 such stations were used to thoroughly test and evaluate the proposed QFT method, as well as to facilitate a direct comparison with more established, classical kernel methods. The selected stations were also classified according to the well-established Köppen climate classification system \cite{koppen_versuch_1900, koppen_handbuch_1930}. \citeauthor{koppen_handbuch_1930} defined five principal global climate zones, which are designated by capital letters: tropical (A), dry (B), mild temperate (C), snow (D), and polar (E). Beyond these primary classifications, the system further specifies subgroups, which divide the world into 31 distinct climatic zones, identifiable through a three-letter nomenclature. The specific set of stations incorporated into this paper, along with their corresponding classifications, are presented and detailed in Table \ref{tab:3}.

\begin{table}[!htb]
\centering
\footnotesize
    \caption{Geographic and climatic details of the 30 BSRN stations.} 
\begin{tabular}{cccccc}
\toprule
\textbf{Code} & \textbf{Station} & \textbf{Lat. ($^{\circ}$N)} & \textbf{Long. ($^{\circ}$E)} & \textbf{Elev. (m)} & \textbf{Climate} \\
\hline
MAN 	&	 Momote 	&	-2.06	&	147.43	&	6 &	 Af \\
BRB 	&	 Brasilia 	&	-15.6	&	-47.71	&	1023	&	 Aw \\
DAR 	&	 Darwin 	&	-12.43	&	130.89	&	30	&	 Aw \\
MNM 	&	 Minamitorishima 	&	24.29	&	153.98	&	7	&	 Aw \\ 
\hline
PTR 	&	 Petrolina 	&	-9.07	&	-40.32	&	387	&	 BSh \\
DAA 	&	 De Aar 	&	-30.67	&	23.99	&	6	&	 BSk \\
SBO 	&	 Sede Boquer 	&	30.86	&	34.78	&	500	&	 BWh \\
GOB 	&	 Gobabeb 	&	-23.56	&	15.04	&	407	&	 BWk \\
\hline
BIL 	&	 Billings 	&	36.61	&	-97.52	&	317	&	 Cfa \\
CLH 	&	 Chesapeake Light 	&	36.91	&	-75.71	&	37	&	 Cfa \\
E13 	&	 Southern Great Plains 	&	36.61	&	-97.48	&	318	&	 Cfa \\
FLO 	&	 Florianopolis 	&	-27.61	&	-48.52	&	11	&	 Cfa \\
FUA 	&	 Fukuoka 	&	33.58	&	130.38	&	3	&	 Cfa \\
ISH 	&	 Ishihakijima 	&	24.34	&	124.16	&	6	&	 Cfa \\
SMS 	&	 Sao Martinho da Serra 	&	-29.44	&	-53.82	&	489	&	 Cfa \\
TAT 	&	 Tateno 	&	36.06	&	140.13	&	25	&	 Cfa \\
BOU 	&	 Boulder 	&	40.05	&	-105.01	&	1577	&	 Cfb \\
CAB 	&	 Cabauw 	&	51.97	&	4.93	&	0	&	 Cfb \\
CAR 	&	 Carpentras 	&	44.08	&	5.06	&	100	&	 Cfb \\
CNR 	&	 Cener 	&	42.82	&	-1.6	&	471	&	 Cfb \\
LIN 	&	 Lindenberg 	&	52.21	&	14.12	&	125	&	 Cfb \\
PAL 	&	 Palaiseau 	&	48.713	&	2.21	&	156	&	 Cfb \\
PAY 	&	 Payerne 	&	46.82	&	6.94	&	491	&	 Cfb \\
IZA 	&	 Izana 	&	28.31	&	-16.5	&	2372	&	 Csb \\
\hline
REG 	&	 Regina 	&	50.21	&	-104.71	&	578	&	 Dfb \\
SAP 	&	 Sapporo 	&	43.06	&	141.33	&	17	&	 Dfb \\
\hline
GVN 	&	 Georg von Neumayer 	&	-70.65	&	-8.25	&	42	&	 EF \\
SYO 	&	 Syowa 	&	-69.01	&	39.59	&	18	&	 EF \\
ALE 	&	 Alert 	&	82.49	&	-62.42	&	127	&	 ET \\
NYA 	&	 Ny-Alesund 	&	78.93	&	11.93	&	11	&	 ET \\
\bottomrule
\label{tab:3}
    \end{tabular}
\end{table}

\subsection{"Classical" tools}
\subsubsection{Pre-Processing}

The general goal is to forecast time–series data. Let 
$\mathbb{D}=\{x_{t_1},x_{t_2},\dots,x_{t_{D_1}}\}\subset\mathbb{R}$ 
be a collection of points representing a quantity at $D_1$ equally spaced timesteps.
We split $\mathbb{D}$ into three collections of points: 
$\mathbb{D}_{train}$, $\mathbb{D}_{val}$, and $\mathbb{D}_{test}$, in chronological order.
All three datasets are standardized using parameters fit on $\mathbb{D}_{train}$ only and then applied to
$\mathbb{D}_{val}$ and $\mathbb{D}_{test}$.

Taking inspiration from classical machine learning, windowing techniques are used to transform
$\mathbb{D}_{train}$, $\mathbb{D}_{val}$, and $\mathbb{D}_{test}$ into collections of windows.
Let $W$ denote the window size and $s\in\mathbb{N}^*$ the stride.
For a forecast horizon $H\in\mathbb{N}^*$ (set $H=1$ for one–step–ahead), each window 
$W^{(i)}$ is paired with the target $x_{t_{l_i+W-1+H}}$ where $l_i \in \{1,2,\dots D_1\}$.

Windows are constructed inside each split only and do not cross split boundaries.
In particular, $\mathbb{D}_{val}$ only contains windows that are chronologically after the windows in $\mathbb{D}_{train}$,
and $\mathbb{D}_{test}$ only contains windows that are chronologically after those in $\mathbb{D}_{val}$.
To ensure there is no data leakage, the last window of one dataset and the first window of the next share no timesteps, such that:
\\
\\
\begin{equation}
 \small
    \mathbb{D}_{test} = \Biggl\{ W^{(1)} = \begin{bmatrix}
           x_{t_{l_1}} \\
           x_{t_{l_1+1}} \\
           \vdots \\
           x_{t_{l_1+W-1}}
         \end{bmatrix}, W^{(2)} = \begin{bmatrix}
           x_{t_{l_2}} \\
           x_{t_{l_2+1}} \\
           \vdots \\
           x_{t_{l_2+W-1}}
         \end{bmatrix}, \dots , W^{(N_{test})} =\begin{bmatrix}
           x_{t_{l_{N_{test}}}} \\
           x_{t_{l_{N_{test}}+1}} \\
           \vdots \\
           x_{t_{l_{N_{test}}+W-1}}
         \end{bmatrix} \Biggr\}
\end{equation} \label{dtest}
\\
where $N_{test}$ is the number of windows, $W$ the window size and $l_i \in \{1,2,\dots D_1\}$, $i = 1, 2, \dots , N_{test}$, $l_{i+1} = s + l_i$ where $s \in \mathbb{N}^*$ is the stride. $\mathbb{D}_{train}$ and $\mathbb{D}_{val}$ are constructed in a similar way and contains $N_{train}$ and $N_{val}$ windows respectively.

\subsubsection{Regression} 
         
Consider window  $W^{(i)} \in \mathbb{D}_{test}$, the goal is to predict the value $x_{t_{l_i+W}}$ ($H=1$). Let $\bold{y}_{train}$, $\bold{y}_{val}$ and $\bold{y}_{test}$ be vectors containing all the target values (for all windows) at future time steps:

\begin{equation}
\bold{y}_{test} = [ x_{t_{l_1+W}}, x_{t_{l_2+W}}, \dots , x_{t_{l_{N_{test}}+W}} ]
\end{equation}

\noindent $\bold{y}_{train}$ and $\bold{y}_{val}$ are constructed in the same way. Using kernel ridge regression, one can compute the predicted value of $x_{t_{l_i+W}}$: 

\begin{equation}
\Tilde{x}_{t_{l_i+W}} = \sum^{N_{train}}_{j=1} \hat{\alpha}_jk(\Tilde{W}^{(j)},W^{(i)})
\end{equation}

\noindent where $i = 1,2,\dots , N_{test}$, $\hat{\bold{\alpha}} = argmin_{\bold{\alpha} \in \mathbb{R}^{N_{train}}} \lVert \bold{y}_{train} - K\bold{\alpha} \rVert^2_2 + \lambda \alpha^TK\alpha$, $\lambda$ is a tunable parameter, $K$ is the kernel matrix such that $K_{uv} = k(\Tilde{W}^{(u)}, \Tilde{W}^{(v)})$, $u, v = 1, 2, \dots , N_{train} $, $\Tilde{W}^{(u)} \in \mathbb{D}_{train}$. The train-test kernel matrix is defined as $\hat{K}_{ij} = k(\Tilde{W}^{(j)},W^{(i)})$, $j = 1, 2, \dots , N_{train} $, $i = 1, 2, \dots , N_{test} $,  $\Tilde{W}^{(j)} \in \mathbb{D}_{train}$ and $\Tilde{W}^{(i)} \in \mathbb{D}_{test}$ so that $\Tilde{x}_{t_{l_i+W}} = \sum^{N_{train}}_{j=1} \hat{\alpha}_j\hat{K}_{ij}$. The kernel matrices that will be used for the regressor will be computed with the QFT based kernel. 

\subsection{"Quantum" tools}


\subsubsection{Quantum kernels}
Presented in \citeauthor{schuld_supervised_2021} \cite{schuld_supervised_2021}, quantum kernels are the quantum analogue of classical kernels, tools that are used mainly in classical machine learning. They are a similarity measure between two datapoints that is computed by taking the inner product of the two datapoints mapped to a higher dimensional space (feature space). These quantum kernels are used mostly for classification tasks using techniques such as support vector Machine and kernelized ridge regression. By using the right feature space, the datapoints in the feature space are "more" linearly separable, making the classification task easier.

In quantum physics/computing, the transformation of a state in a closed system is implemented by a unitary operator $U$ satisfying $UU^\dagger =\mathbb{I}$, so norms and inner products (hence probabilities) are preserved. In order to embed a classical data point $x \in \mathbb{C}^m$ in a point in the quantum feature space $\ket{\phi(x)} \in \mathbb{C}^k$, $m,k \in \mathbb{N}^*$, it can be define a quantum embedding $U(x)$ that completely determines the feature space. The equality $\ket{\phi(x)} = U(x)\ket{0}$ is then obtained. Thus, it can define the kernel value between two datapoints $x$ and $y$:

\begin{align}
\begin{split}
    k(y,x) &= |\bra{0}U^{\dagger}(y)U(x)\ket{0}|^2  \\
    &=  |\braket{\phi(y)|\phi(x)}|^2 \\
    &=  k(x,y)
    \end{split}
\end{align}

Such a quantum kernel can then be implemented on a quantum computer by successively applying the unitaries $U(x)$ and $U^{\dagger}(y)$ to the initial state $\ket{0}$.

\begin{figure}[H]
\begin{center}
   \begin{quantikz}
\ket{0}&\gate[wires=4, nwires=3]{U(x)}&[2cm]\gate[wires=4, nwires=3]{U^\dagger(y)}&[2cm] \meter{} \\
\ket{0}&& & \meter{} \\
 \vdots & \vdots &&  \vdots \\
\ket{0}&& &[2cm]\meter{}
\end{quantikz}
\end{center}
\caption{General quantum circuit used to compute the kernel value between the two datapoints x and y.} \label{fig:scheme3}
\end{figure}
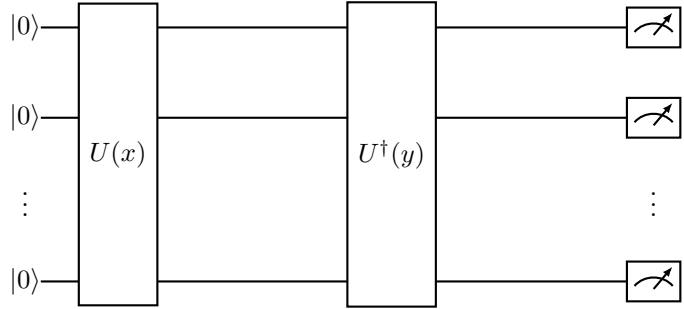

Using this circuit of $n$ qubits, the probability of measuring the state $\ket{0}^{\otimes n}$ corresponds to the kernel value $k(x,y)$:

\begin{align}
\begin{split}  
    P(\ket{0}^{\otimes n})  &= |\bra{0}^{\otimes n}U^{\dagger}(y)U(x)\ket{0}^{\otimes n}|^2  \\
    &=  k(x,y)
\end{split}
\end{align}

This is how kernel values are obtained using a quantum computer\footnote{\noindent There exists other Quantum circuits architecture to compute kernel values, such as Projected kernels, but this study focuses on the type of circuits presented above. }. There are multiple existing embeddings in the literature, such as Amplitude encoding, which encodes a classical datapoint $x \in \mathbb{C}^{2^n}$ of size $2^n$ into n qubits (the values of x are encoded in the amplitudes of the basis states): 

\begin{equation}
x = \begin{bmatrix}
           x_0 \\
           x_1 \\
           \vdots \\
           x_{2^n -1}
         \end{bmatrix}  \mapsto \ket{x} = \sum_{i=0}^{2^n-1} x_i \ket{i}    
\end{equation}






\subsubsection{Quantum Fourier Transform (QFT)}

Presented by \citeauthor{coppersmith_approximate_2002} \cite{coppersmith_approximate_2002}, the Quantum Fourier Transform (QFT) is the quantum analogue of the discrete Fourier transform. It operates on a quantum bit register to convert the quantum information from the time domain to the frequency domain. This transformation is crucial for many quantum algorithms, offering a speed-up over classical counterparts. The QFT on a register of $n$ qubits is defined as the mapping:

\begin{equation}
\ket{j} \mapsto \frac{1}{\sqrt{2^n}} \sum_{k=0}^{2^n-1} e^{2\pi i j k / 2^n} \ket{k},
\end{equation}

\noindent where $\ket{j}$ and $\ket{k}$ are quantum states of the register. This transformation can be implemented using a series of quantum gates, particularly Hadamard and controlled gates. Suppose that one has an initial state $\ket{x} = \sum_{i=0}^{2^n-1} x_i \ket{i}$ (classical data point x encoded using amplitude encoding) on which QFT is applied, the resulting state is, 

\begin{equation}
QFT\ket{x} = \sum_{k=0}^{2^n-1} y_k \ket{k} 
\end{equation}

\noindent where $y_k = \frac{1}{\sqrt{2^n}} \sum_{j=0}^{2^n-1} x_j e^{2\pi i j k / 2^n}$. In machine learning, especially in time series forecasting, capturing the periodic patterns and understanding the frequency components of past data can be crucial. The QFT, when used within the embedding part of a quantum kernel, could offer an efficient means of analyzing these components.

\subsubsection{QFT based Quantum Kernel}

Embedding time-series data into a quantum state effectively involves mapping classical information (time-series data) into the state of a quantum system. By applying QFT as an embedding step, the data is transformed into the frequency domain, which could allow the quantum kernel to process complex patterns and periodicities inherently present in the time series. This is the main idea that will guide the design of the quantum kernel.

The Discrete Fourier Transform has to be applied to a collection of equally spaced points in time, the same goes for the QFT. Therefore, the input data to the kernel presented here will not be a vector containing different features (which is usually done in the literature), but rather a vector of equally spaced points in time of one given feature. The usual datapoints used by quantum kernels are of the following type:

\begin{equation}
\bold{x} = \begin{bmatrix}
           x_1 \\
           x_2 \\
           \vdots \\
           x_{N_1}
         \end{bmatrix}  
\end{equation}

\noindent where each $x_i$ is the value of a certain feature at a given time ($\bold{x} \in \mathbb{R}^{N_1}$). 

In a forecasting scenario where the first feature is to be predicted, the usual datapoints type (i.e. a vector containing different features) does not work with the QFT idea presented earlier. Therefore, the input datapoints have to be modified such that they are of the following type:

\begin{equation}
\bold{x_1} =\begin{bmatrix}
           x_1^{(t_1)} \\
           x_1^{(t_2)} \\
           \vdots \\
           x_1^{(t_k)}
         \end{bmatrix}    
\end{equation}

\noindent where the coefficients of the vector are the values of the first feature at $k$ equally spaced points in time ($\bold{x_1} \in \mathbb{R}^{k}$).

It is important to note that, for the QFT to yield the most meaningful results, the input must contain a significant number of equally spaced points in time, which means that $k$ should be relatively large. If angle embedding was used, $k$ qubits would be required, which may be too much depending on $k$, hence why amplitude encoding will be used instead, requiring only $log_2(k)$ qubits, thus much more feasible on the current generation of quantum simulators. The data windows will then be compared to each other by the quantum kernel and the results will be used to perform forecasting tasks. The main idea is that 'similar' windows should have 'similar' following values (at the next timestep).
\newline

The quantum embedding part of the non-parameterized QFT based quantum kernel $U(x)$ is divided into three parts:

\begin{itemize}
    \item $A(x)$ : amplitude encoding of x
    \item $QFT$ : Quantum Fourier Transform
    \item $V(x)$ : protective layer
\end{itemize}
The 'protective layer' is there to ensure that $QFT$ and $QFT^\dagger$ don't cancel each other, this also means that it must depend on the input data. It's construction will be discussed later.
\\
\\
Combining these three parts, the following QFT based quantum kernel can be constructed.

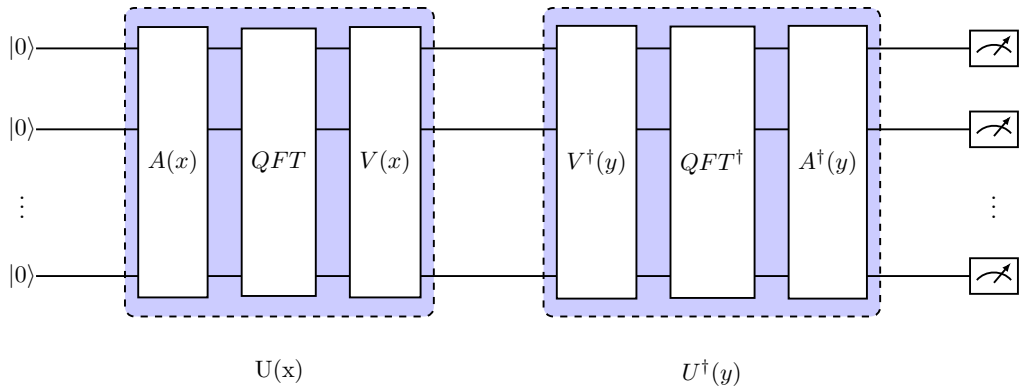
\begin{figure}[H]
\begin{center}
\begin{adjustbox}{width=1\textwidth}
   \begin{quantikz}
\ket{0}&[1cm]\gate[wires=4, nwires=3]{A(x)}\gategroup[4,steps=3,style={dashed,
rounded corners,fill=blue!20, inner xsep=2pt},
background,label style={label position=below,anchor=
north,yshift=-0.7cm}]{U(x)}&\gate[wires=4, nwires=3]{QFT}&\gate[wires=4, nwires=3]{V(x)}&[1.5cm]\gate[wires=4, nwires=3]{V^\dagger(y)}\gategroup[4,steps=3,style={dashed,
rounded corners,fill=blue!20, inner xsep=2pt},
background,label style={label position=below,anchor=
north,yshift=-0.7cm}]{$U^\dagger(y)$}&\gate[wires=4, nwires=3]{QFT^\dagger}&\gate[wires=4, nwires=3]{A^\dagger(y)}&[1cm] \meter{} \\
\ket{0}&&&&&& & \meter{} \\
 \vdots &&& \vdots &&&&  \vdots \\
\ket{0}&&&&&& &[2cm]\meter{}
\end{quantikz}
\end{adjustbox}
\end{center}
\caption{QFT based quantum kernel using n qubits.}\label{fig:QFTkernel}
\end{figure}

In Figure \ref{fig:QFTkernel}, the significance of the protective layer $V(\cdot)$ can be observed. Indeed, without it, $QFT$ and $QFT^\dagger$ would cancel each other out and only an amplitude-encoding quantum kernel will be left. Note that the window size must be a power of 2 as amplitude encoding encodes a vector of size $2^n$ into n qubits. 
\\
\\
There are multiple options for the choice of $V(\cdot)$:

\begin{itemize}
    \item Amplitude encoding
    \item Variation of the angle/rotational encoding
\end{itemize}

Since amplitude encoding is already used ($A(x)$ in Figure \ref{fig:QFTkernel}), encoding $2^n$ values into $n$ qubits, regular angle encoding cannot be used as it would in this case require $2^n$ qubits. Amplitude encoding could be used but for computational speed reasons, a variation of the angle encoding is be used, which encompasses alternating $R_X(\cdot)$ and $R_Y(\cdot)$ gates. Assume a system consisting of three qubits, windows must be of size $2^3 = 8$. The Euclidean division of 8 by 3 is $8 = 2 * 3 + 2$, meaning that there will be 2 rotational gates per qubit, except for the last qubit which will have 4 rotational gates. Here is an example of this embedding for a vector $x = \begin{bmatrix}
           x_1 \\
           x_2 \\
           \vdots \\
           x_8
         \end{bmatrix} \in \mathbb{R}^{8}$.

\begin{figure}[H]
\begin{center}
\begin{adjustbox}{width=1\textwidth}
   \begin{quantikz}
& \gate[wires=3]{V(x)} & \qw \\
& & \qw \\
& & \qw
\end{quantikz}
\hspace{20pt}= \hspace{20pt} \begin{quantikz}
& \gate{R_X(x_1)} & \gate{R_Y(x_2)}  &       \qw          &    \qw           &\qw \\
& \gate{R_X(x_3)} & \gate{R_Y(x_4)}  &      \qw           &    \qw            & \qw \\
& \gate{R_X(x_5)} & \gate{R_Y(x_6)}   &\gate{R_X(x_7)} & \gate{R_Y(x_8)}& \qw
\end{quantikz}
\end{adjustbox}
\end{center}
\caption{Example of the protective layer $V(\cdot)$ on 3 qubits}\label{fig:toto}
\end{figure}
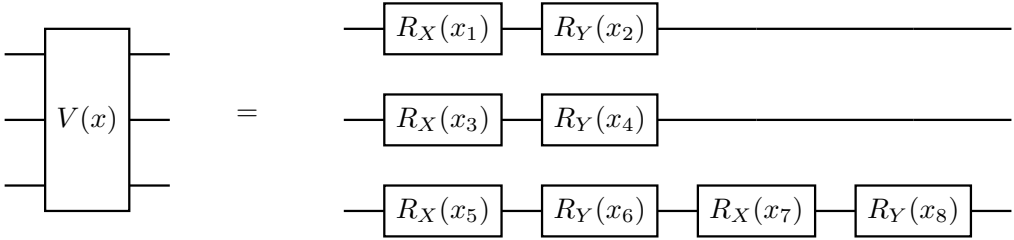

\begin{figure}[H]
\begin{center}
\begin{adjustbox}{width=1\textwidth}
   \begin{quantikz}
& \gate[wires=4, nwires=3]{V(x)} & \qw \\
& & \qw \\
 \vdots & &   \\
& & \qw
\end{quantikz}
\hspace{10pt}= \hspace{10pt} \begin{quantikz}
& \gate{R_X(x_1)}  & \gate{R_Y(x_2)}  & \qw \dots &\gate{R_X(x_{a})} &\qw \\
& \gate{R_X(x_{a+1})} & \gate{R_Y(x_{a+2})}& \qw \dots &\gate{R_X(x_{2a})}& \qw \\
  & & \vdots & & \\
& \gate{R_X(x_{(n-1)a + 1})} & \gate{R_Y(x_{(n-1)a + 2})}& \qw \dots &\gate{R_X(x_{na + b})} & \qw 
\end{quantikz}
\end{adjustbox}
\end{center}
\caption{Protective layer $V(\cdot)$ on n qubits where $a+b$ is odd}\label{fig:yeye}
\end{figure}
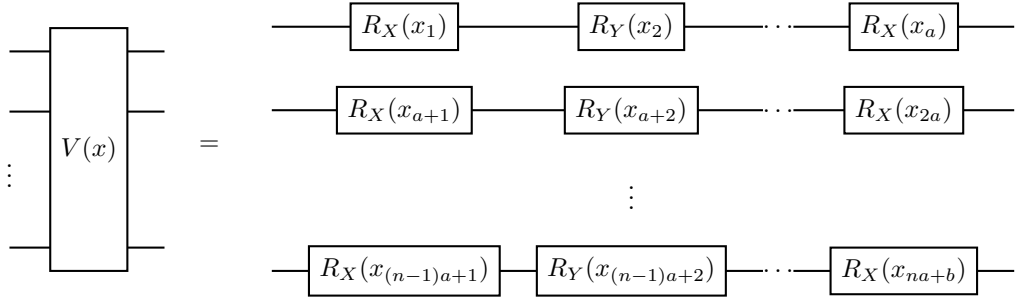

For the more general case where one has n qubits, the Euclidean division of $2^n $ by n, $2^n = an+b$ ($a,b \in \mathbb{N}$) is computed. Suppose that $a+b$ is odd, the last gate of the last qubit is $R_X(\cdot)$ (if $a+b$ is even, the last gate is $R_Y(\cdot)$). For the first $n-1$ qubits, if $a$ is odd, the last gate (for each qubit) is $R_X(\cdot)$ (if a is even, the last gate is $R_Y(\cdot)$). The number of rotational gates on the last qubit is always greater than or equal to the number of rotational gates on the other qubits (for the protective layer). Here is an example of this embedding for a vector $x = \begin{bmatrix}
           x_1 \\
           x_2 \\
           \vdots \\
           x_{2^n}
         \end{bmatrix} \in \mathbb{R}^{2^n}$, as shown in Figure \ref{fig:yeye}.

As developed in \ref{appA}, the kernel value between two windows (considered mathematically as vectors) can be expressed as follows:

\begin{equation}
    k(W_s^{(i)}, W_s^{(j)}) = |\,\mathrm{Tr}(\sigma^{(i,j)}_s R^{(i,j)}_s)\,|^2
\end{equation}

\noindent
where $W_s^{(i)}$ represents the $i$-th window corresponding to feature $s$, $N = 2^n$ (with $n$ qubits), and matrices $\sigma^{(i,j)}_s$ and $R^{(i,j)}_s$ are defined by:
\begin{align*}
    \sigma^{(i,j)}_s = (\sigma_{pk}^{(i,j,s)}) &= (\tilde{y}_p^{(i,s)}\,y_k^{(j,s)}), \\
    R^{(i,j)}_s &= V^\dagger(W_s^{(i)})\,V(W_s^{(j)}),
\end{align*}
with
\begin{align*}
    y^{(j,s)}_k &= \frac{1}{\sqrt{N}} \sum_{v=0}^{N-1}(W_s^{(j)})_v \, e^{2\pi i \frac{vk}{N}}, \\
    \tilde{y}_p^{(i,s)} &= \frac{1}{\sqrt{N}} \sum_{v=0}^{N-1}(W_s^{(i)})_v \, e^{-2\pi i \frac{vp}{N}}.
\end{align*}

Through further development presented in \ref{appB}, the following equality is obtained:
\begin{equation}
    k(W_s^{(i)}, W_s^{(j)}) = |\frac{1}{N} \sum_{l, v} \beta_{lv}^{(i,j,s)} Tr(\Omega^{(l,v)}R^{(i,j)}_s)|^2
\end{equation}
\noindent where $\beta_{lv}^{(i,j,s)} =(W_s^{(i)})_l (W_s^{(j)})_v $, $\Omega^{(l,v)} = \begin{bmatrix}
    1 & \omega_N^{v} & \dots  & \omega_N^{v(N-1)} \\
    \omega_N^{-l} & \omega_N^{v-l} &   &  \\
    \vdots &   & \ddots & \vdots \\
    \omega_N^{-l(N-1)} &  & \dots  & \omega_N^{(v-l)(N-1)}
\end{bmatrix}$, \\$\omega_N = e^{\frac{2 \pi i }{N}}$
\newline
%
%


Moreover, in \ref{appC} it is developed the value of $R^{(i,j)}_s$ in the case of the rotational encoding variation.

\subsubsection{Feature combination and selection} 

The feature combination technique developed here is similar to the kernel combination technique presented by \citeauthor{baker_parallel_2024} \cite{baker_parallel_2024}. 

The idea is that for each additional feature, the data will be transformed into a collection of windows which all are at the same time steps as the original feature windows (the same train/test split is used). All features must share a common time axis. For each feature $s$, ($s=1,2,\dots,M$), the kernel matrices $K_s$ (train-train) and $\hat{K}_s$ (test-train) presented earlier will be computed. These matrices will then be combined separately using a weighted sum to obtain the two final kernel matrices $K^{(f)} = \sum^{M}_{s=1} \beta_s K_s$ and $\hat{K}^{(f)} =  \sum^{M}_{s=1}\beta_s \hat{K}_s$ containing both feature-related and temporal information.

\begin{figure}[H]
    \centering
    \includegraphics[width=0.85\textwidth]{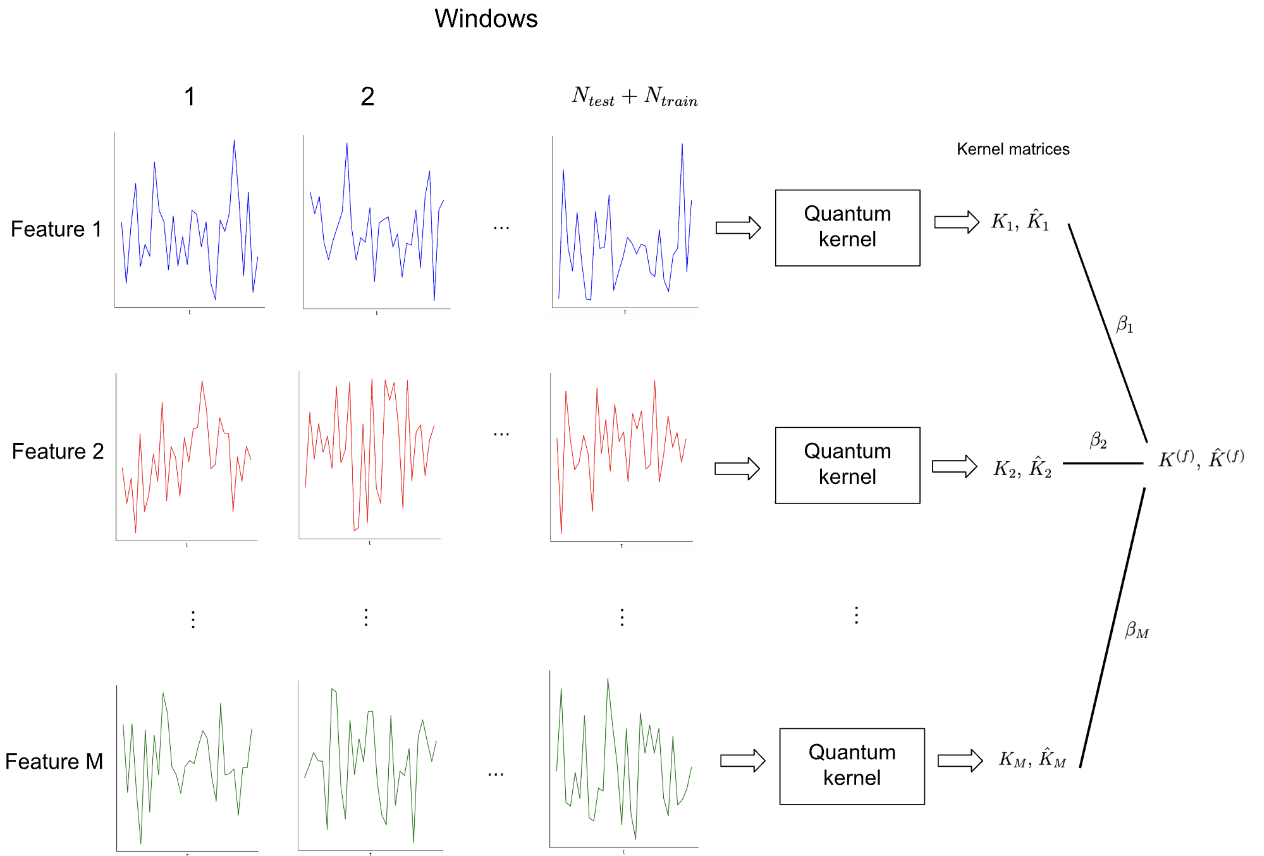}
    \caption{Feature combination scheme using toy datasets.}
    \label{fig:schemefeatures}
\end{figure}

One way one can look at these new kernel matrices is as a collection of similarity measures between collections of windows. Let $W^{(i)}_s$ be the i-th testing window of the s-th feature (similarly, $\Tilde{W}^{(j)}_s$ is the j-th training window of the s-th feature), two multivariate time series datapoints can now defined as 

\begin{equation}
X_i=
    \begin{bmatrix}
            W_1^{(i)}\\
            W_2^{(i)} \\
            \vdots \\
            W_M^{(i)}
    \end{bmatrix}
, \Tilde{X}_j 
= 
\begin{bmatrix}
            \Tilde{W}_1^{(j)}\\
            \Tilde{W}_2^{(j)} \\
            \vdots \\
            \Tilde{W}_M^{(j)}
         \end{bmatrix}
\end{equation}

\noindent such that $k^{(f)}(\tilde{X}_i, \tilde{X}_j) = \sum^{M}_{s = 1} \beta_s k(\tilde{W}_s^{(i)},\tilde{W}_s^{(j)}) = \sum^{M}_{s = 1} \beta_s K_{ij} = K^{(f)}_{ij}$ and $\hat{k}^{(f)}(X_i, \tilde{X}_j) = \sum^{M}_{s = 1} \beta_s k(W_s^{(i)},\tilde{W}_s^{(j)}) = \sum^{M}_{s = 1} \beta_s \hat{K}_{ij} = \hat{K}^{(f)}_{ij}$. $K^{(f)}$ and $\hat{K}^{(f)}$ are then used by the kernel ridge regressor so that it is trained to predict $\bold{y}_{train}$. It can then generate a prediction $\tilde{\bold{y}}_{test}$ of $\bold{y}_{test}$. Figure \ref{fig:schemefeatures} shows the schema for combining the feature.

In the context of solar radiation, the features were selected by a lag-1 correlation screen on the training split with the future target ($\mathrm{Glo\_rad}_{t+1}$, the time change was only applied for the correlation computations), complemented by physical considerations of solar geometry and clear sky irradiance. The final set comprises global irradiance (\texttt{Glo\_rad}, the target feature), hour angle (\texttt{Hour\_angle}), solar altitude (\texttt{Solar\_altitud}), and clear-sky radiative components (\texttt{Direct\_normal\_clear\_sky\_rad}, 
\\ 
\texttt{Global\_clear\_sky\_rad}, \texttt{Diffuse\_clear\_sky\_rad}). All inputs are standardized per station and windowed with $L=32$ steps (thus $n=5$ qubits for the quantum branch).

\subsubsection{Classical kernels for comparison}

To benchmark the performance of the quantum Fourier transform (QFT)-based quantum kernel, two classical kernels widely used in machine learning are considered: the radial basis function (RBF) kernel and the polynomial kernel \cite{RBFPOLY}. Each input to these kernels consists of two vectors, $\mathbf{x}$ and $\mathbf{x}'$, representing a pair of temporal windows from the original time series data, constructed identically to the quantum kernel inputs.

The RBF kernel is defined as:
\begin{equation}
k_{\mathrm{RBF}}(\mathbf{x},\mathbf{x}')=\exp(-\gamma||\mathbf{x} - \mathbf{x}'||^{2})
\end{equation}
where $\gamma$ controls the sensitivity to the squared Euclidean distance between the vectors $\mathbf{x}$ and $\mathbf{x}'$.

The polynomial kernel is defined as:
\begin{equation}
k_{\mathrm{Poly}}(\mathbf{x},\mathbf{x}')=(\gamma \mathbf{x}^\top \mathbf{x}' + r)^d
\end{equation}
where $\gamma$ scales the dot product, $r$ is a constant offset, and $d$ determines the polynomial degree, capturing global non-linear relationships.

Both kernels are computed separately for each feature and each pair of time windows, using identical train-validation-test splits as those applied in the quantum kernel experiments. After computing the kernels, classical kernel combinations are optimized through Bayesian optimization \cite{snoek2012practical} to select their optimal weights and ridge regularization parameters, as described in detail in Section~\ref{Opti}. This approach ensures a rigorous and fair comparison of the quantum kernel's performance against established classical kernel methods.

We keep the classical kernel hyperparameters fixed (e.g., RBF bandwidth; polynomial degree, scale, and offset) and do not tune them. The only parameters we tune for both (classical and quantum) models are the convex feature matrices-mixing weights and the KRR ridge parameter.



\subsubsection{Optimization} \label{Opti}

For every station, the goal is to determine  
(i) a convex weight vector
\(w=(w_{1},\dots ,w_{k})\) that combines the \(k\) feature-specific
kernels into
\(K_{\mathrm{mix}}(w)=\sum_{f=1}^{k}w_{f}K^{(f)}\) and  
(ii) the ridge coefficient \(\alpha>0\) of kernel-ridge regression
(KRR), so as to \textit{maximise the coefficient of determination
\(R^{2}\) on the fixed validation split}.
Identical train–validation–test partitions and the same metric
(\(R^{2}\)) are used for classical and quantum kernels.

\subsubsection*{Optimisation algorithm}

\begin{enumerate}[label=\textbf{Step \arabic*.}, leftmargin=1.47cm]
\item \textbf{Bayesian proposal of a weight vector.}  
      Each feature-specific kernel \(K^{(f)}\) receives a non-negative
      weight \(w_{f}\) that determines its contribution to the mixed
      kernel  
      \(K_{\mathrm{mix}}(w)=\sum_{f=1}^{k}w_{f}K^{(f)}\).
      Because the optimiser can only handle box constraints, the weights
      are generated in two ways:

      \begin{itemize}[leftmargin=0.6cm]
      \item Classical branch.\;
            The optimiser first proposes an unconstrained latent
            vector \(v\in[-4,4]^{k}\).
            A soft-max transformation  
            \[
              w_{f}
              =\frac{\exp(v_{f})}{\sum_{j=1}^{k}\exp(v_{j})}
            \]
            maps \(v\) to a valid weight vector, guaranteeing
            \(w_{f}>0\) and \(\sum_{f=1}^{k}w_{f}=1\).

      \item Quantum branch.\;
            The optimiser proposes a raw vector
            \(\tilde{w}\in[0,1]^{k}\).
            Simple renormalisation  
            \(w_{f}=\tilde{w}_{f}/\sum_{j=1}^{k}\tilde{w}_{j}\)
            enforces the same positivity and unity-sum constraints.
      \end{itemize}

\item \textbf{Inner grid search for \(\alpha\).}  
      For the fixed mixture \(K_{\mathrm{mix}}(w)\) a one-dimensional
      grid search is carried out over a large number of values for $\alpha$,
      fitting KRR and recording \(R^{2}_{\mathrm{Validation}}\) at each grid
      point.  The highest value is denoted
      \(R^{2}_{\max}(w)\).

\item \textbf{Return value to the optimiser.}  
      The optimiser receives
      \(-R^{2}_{\max}(w)\) as the scalar objective, updates its
      surrogate model, and proposes the next weight vector.
      Iteration continues until the 20-call limit is exhausted.

\item \textbf{Numerical safeguard (classical branch only).}  
      Before every KRR fit, a diagonal jitter
      \[
          \varepsilon=
          10^{-6}\,
          \frac{\operatorname{tr}\!\bigl(K_{\mathrm{mix}}(w)\bigr)}{n}
      \]
      is added :
      \(K_{\mathrm{mix}}\leftarrow K_{\mathrm{mix}}+\varepsilon I\),
      guaranteeing positive-definiteness and avoiding failures.

\end{enumerate}

\subsubsection*{Final model and reporting}

The weight vector \(w^{\star}\) and ridge parameter \(\alpha^{\star}\)
that maximise \(R^{2}_{\mathrm{Validation}}\) are retained and used on the test set.

\textit{In essence:}  
“propose \(w\)\(\rightarrow\) scan 100 $\alpha$ values
\(\rightarrow\) return the best \(-R^{2}_{\mathrm{Validation}}\)
\(\rightarrow\) refine \(w\) via Bayesian optimisation.”
Soft-max and jitter appear only in the classical branch to ensure
numerical stability; they do not alter the optimisation objective,
thereby preserving comparability with the quantum results.

\bigskip
\noindent



\subsection{Statistical indicators}  

To assess the performance of the QML model and conventional ML models in forecasting the global horizontal irradiance for the subsequent hour $t+1$, three statistical metrics are employed: $nRMSE$, $nMBE$, and $R^2$.

\begin{equation}\label{eq:4.9}
    nRMSE = \frac{\sqrt{\frac{1}{N}\sum_{i=1}^{N} (x - y)^2 }}{\bar{y}} \times 100
\end{equation}

The normalized root mean square error ($nRMSE$), represented by Eq.~(\ref{eq:4.9}), reflects how closely a model's predictions ($x$) align with actual observations ($y$). Specifically, it captures the root of the averaged squared differences between estimates and measurements, then scales that quantity by the mean of the observed values ($\bar{y}$) and expresses the result as a percentage. Lower values of $nRMSE$ suggest that, on average, the model's predictions are nearer to the observed data.

\begin{equation}\label{eq:4.10}
    nMBE = \frac{\frac{1}{N} \sum_{i=1}^{N} (x - y)}{\bar{y}} \times 100
\end{equation}

Meanwhile, the normalized mean bias error ($nMBE$), as shown in Eq.~(\ref{eq:4.10}), offers insight into the overall bias of the predictive model. This metric aggregates the individual discrepancies ($x - y$) over all $N$ samples, computes the mean of those discrepancies, and then normalizes that mean by the average observed value ($\bar{y}$) before converting it to a percentage. An $nMBE$ value approaching zero indicates that, across the entire dataset, the model introduces little systematic bias in its predictions.

\begin{equation}\label{eq:4.11}
    R^{2} = \frac{\bigl(\sum_{i=1}^{N}(x - \bar{x})(y - \bar{y}) \bigr)^{2}}
    {\sum_{i=1}^{N}(x - \bar{x})^{2}\,\sum_{i=1}^{N}(y - \bar{y})^{2}}
\end{equation}

Lastly, the coefficient of determination ($R^{2}$), defined by Eq.~(\ref{eq:4.11}), quantifies the proportion of variance in the observed data ($y$) that is accounted for by the model's predictions ($x$). It is computed by squaring the sum of cross-deviations between $x$ and $y$, and dividing that value by the product of the sums of their respective squared deviations from their means. Higher values of $R^{2}$, especially those nearing unity, indicate that the model effectively captures the variability present in the observed measurements.

\section{Results}

\begin{figure}[!htb]
    \centering
    \includegraphics[width=0.9\linewidth]{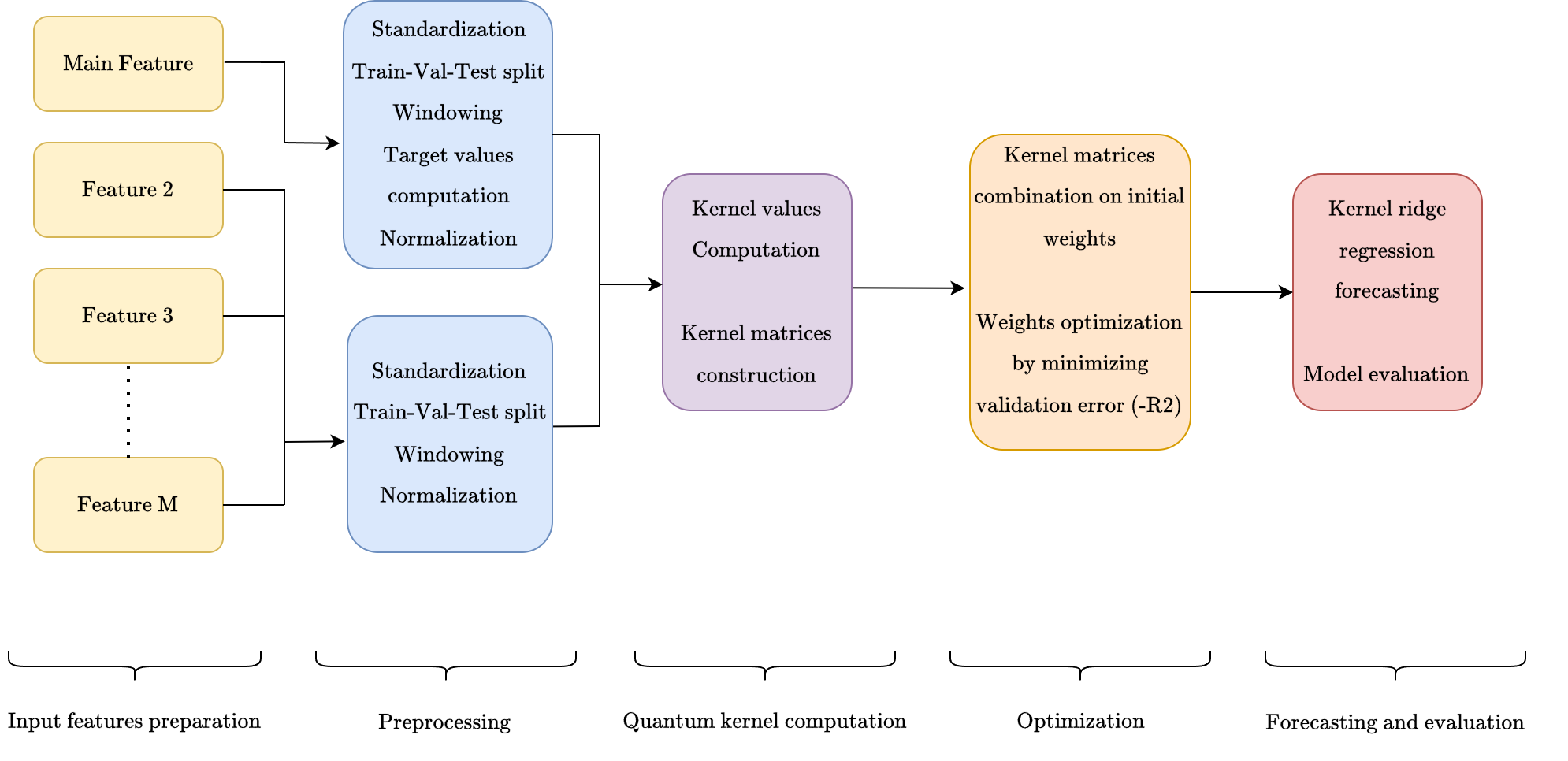}
    \caption{Workflow of the study: feature preprocessing, per-feature kernel construction, kernel fusion via weight optimisation, and kernel-ridge forecasting.}
    \label{fig:pipeline}
\end{figure}

All quantum kernel computations reported in this study were performed using the \texttt{PennyLane} library with the noiseless state-vector simulator \texttt{lightning.qubit}, simulating quantum circuits of 5 qubits (corresponding to windows of size $2^5 = 32$). On average, each station supplied 1950 training windows, 242 validation windows and  242 test windows ($\approx$ 80 \% / 10 \% / 10 \% split). The methodological workflow that precedes the empirical analysis is summarised in Figure~\ref{fig:pipeline}. First, each raw time series variable including the response (hereafter, \emph{main feature}) and the remaining \(M-1\) auxiliary predictors was processed separately: (i) standardisation to zero mean and unit variance; (ii) chronological partitioning into training, validation, and test subsets; (iii) conversion into overlapping windows of fixed length \(W\); and (iv) extraction of a one step ahead target value for the main feature so that the forecasting task becomes \(W^{(i)} \mapsto x_{t_{i}+W}\). Finally, all windows were individually normalized (L2 normalization) to be encoded into the quantum circuit.

For every feature, three kernel families were computed: the proposed QFT-based quantum kernel and two classical references (radial–basis-function and polynomial). This produced per–feature train–train and test–train kernel matrices that were fused separately through convex linear combinations whose weights were optimised on the validation sets by minimising the negative coefficient of determination, \(-R^{2}\). The optimised composite kernels served as inputs to a kernel–ridge regressor, whose ridge coefficient was likewise selected on the validation data. The resulting models were finally applied to the unseen test windows, and their forecasts were assessed with the normalised root-mean-square error (\(n\mathrm{RMSE}\)), the normalised mean bias error (\(n\mathrm{MBE}\)), and the coefficient of determination (\(R^{2}\)), as reported in the following sections.

\begin{figure}[!htb]
    \centering
    \includegraphics[width=0.8\linewidth]{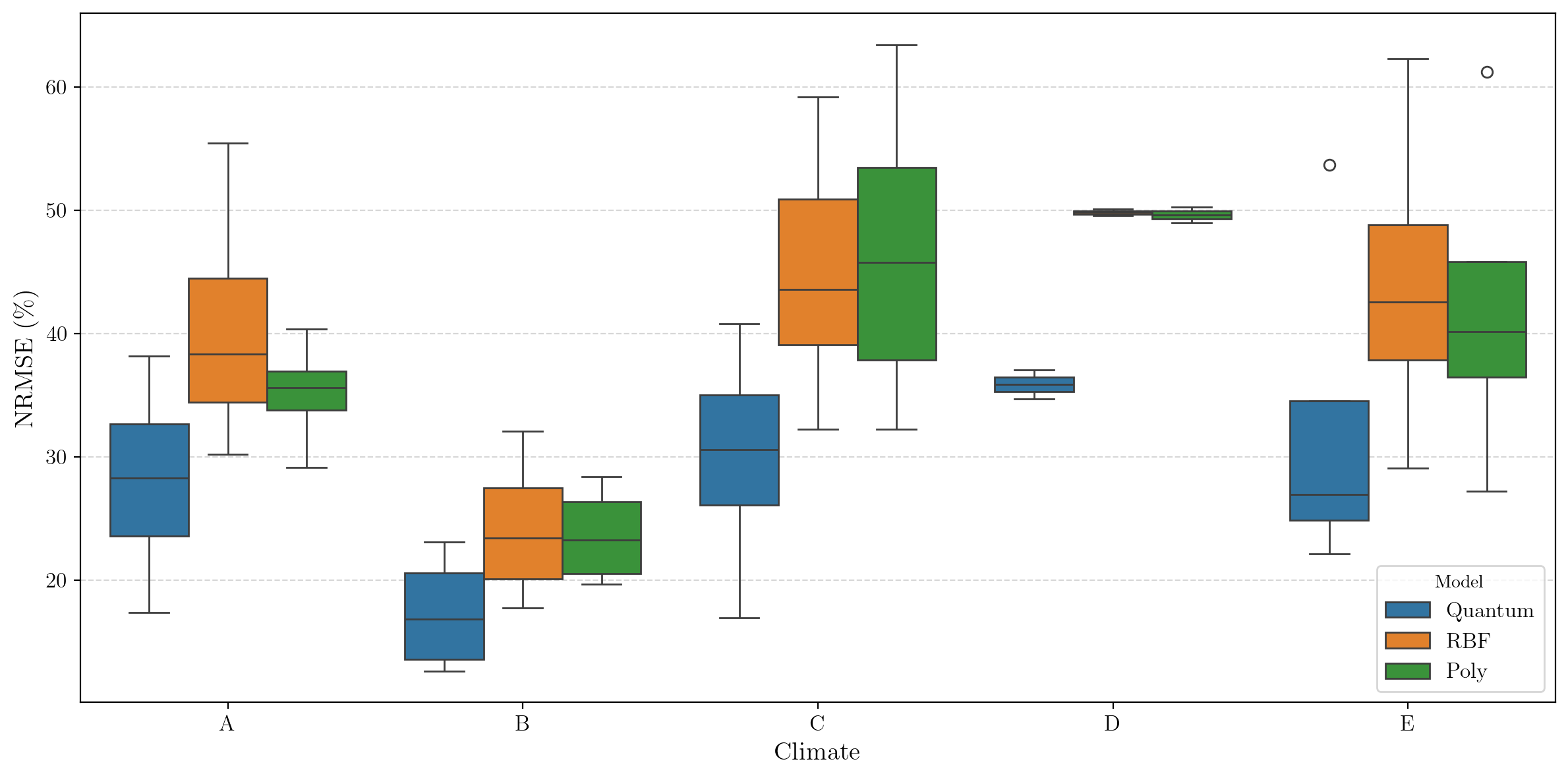}
    \caption{Normalized root‐mean‐square error (NRMSE) for the quantum kernel (QK) and two classical kernels—radial basis function (RBF) and polynomial (Poly)—grouped by Köppen climate class.}
    \label{fig:1}
\end{figure}

The distributions of NRMSE in Figure~\ref{fig:1} show a clear and consistent advantage for the quantum kernel (QK) over the classical benchmarks in every Köppen \& Geiger climate class. In tropical climates (A), the QK median is roughly \(28\,\%\), or about 7–10 percentage points below those of RBF and Poly. This margin widens in arid regions (B), where QK attains a median near \(17\,\%\) versus \(23\text{–}24\,\%\) for the classical kernels. For temperate sites (C) the QK cuts the central error to approximately \(31\,\%\), outperforming RBF and Poly by 13–15 points while suppressing upper‐tail spread. snow stations (D) maintain the same ordering, with QK around \(36\,\%\) and both classical medians clustering near \(50\,\%\). Finally, in polar zones (E) QK records a median near \(27\,\%\), whereas RBF and Poly register \(42\,\%\) and \(40\,\%\), respectively, and exhibit high-error outliers that exceed \(60\,\%\). Taken together, these results demonstrate that the proposed quantum kernel consistently and robustly outperforms its classical counterparts across all five climatic zones.

\begin{figure}[!htb]
    \centering
    \includegraphics[width=0.8\linewidth]{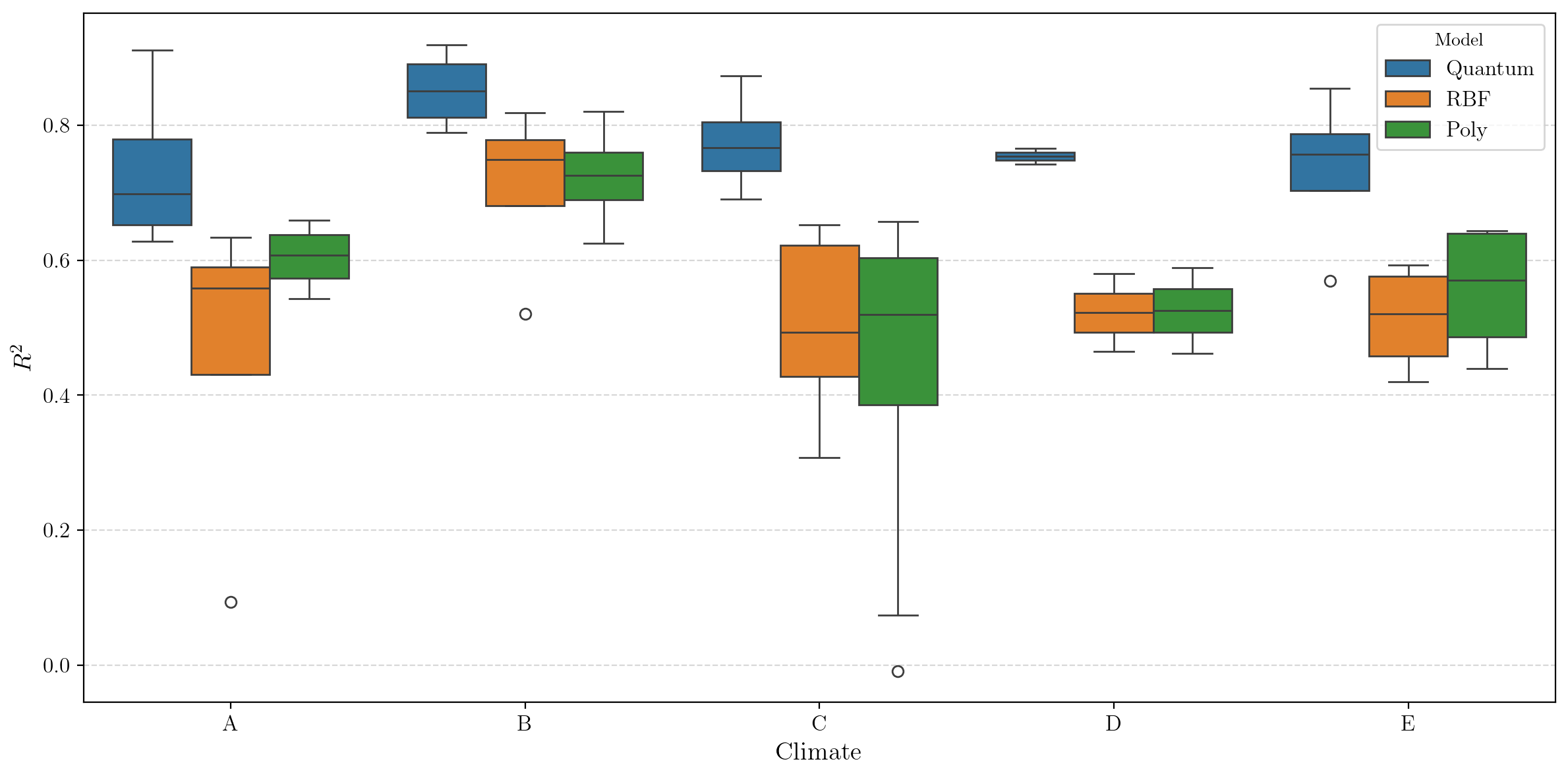}
    \caption{Coefficient of determination (\(R^{2}\)) for QK, RBF, and Poly kernels by Köppen climate class.}
    \label{fig:2}
\end{figure}

Across the same climate classes, Figure~\ref{fig:2} reveals that \(R^{2}\) likewise favors the quantum kernel.  In tropical stations, QK achieves a median close to \(0.72\), exceeding the RBF and Poly medians by roughly 0.15 and 0.10, respectively, while maintaining a tightly bounded IQR despite an extreme RBF outlier below 0.15.  The advantage grows in arid climates, where QK posts a median near \(0.87\) with minimal dispersion of 0.10 above the classical kernels. Temperate sites show QK at \(0.75\), whereas RBF and Poly drop to about \(0.50\); the Poly kernel also exhibits a negative outlier, highlighting instability.  Snow stations again place QK on top with a median of \(0.76\) versus \(0.53\) for both classical models. In polar environments the pattern persists: medians of \(0.75\), \(0.55\), and \(0.64\) for QK, RBF, and Poly, respectively.

\begin{figure}[!htb]
    \centering
    \includegraphics[width=0.8\linewidth]{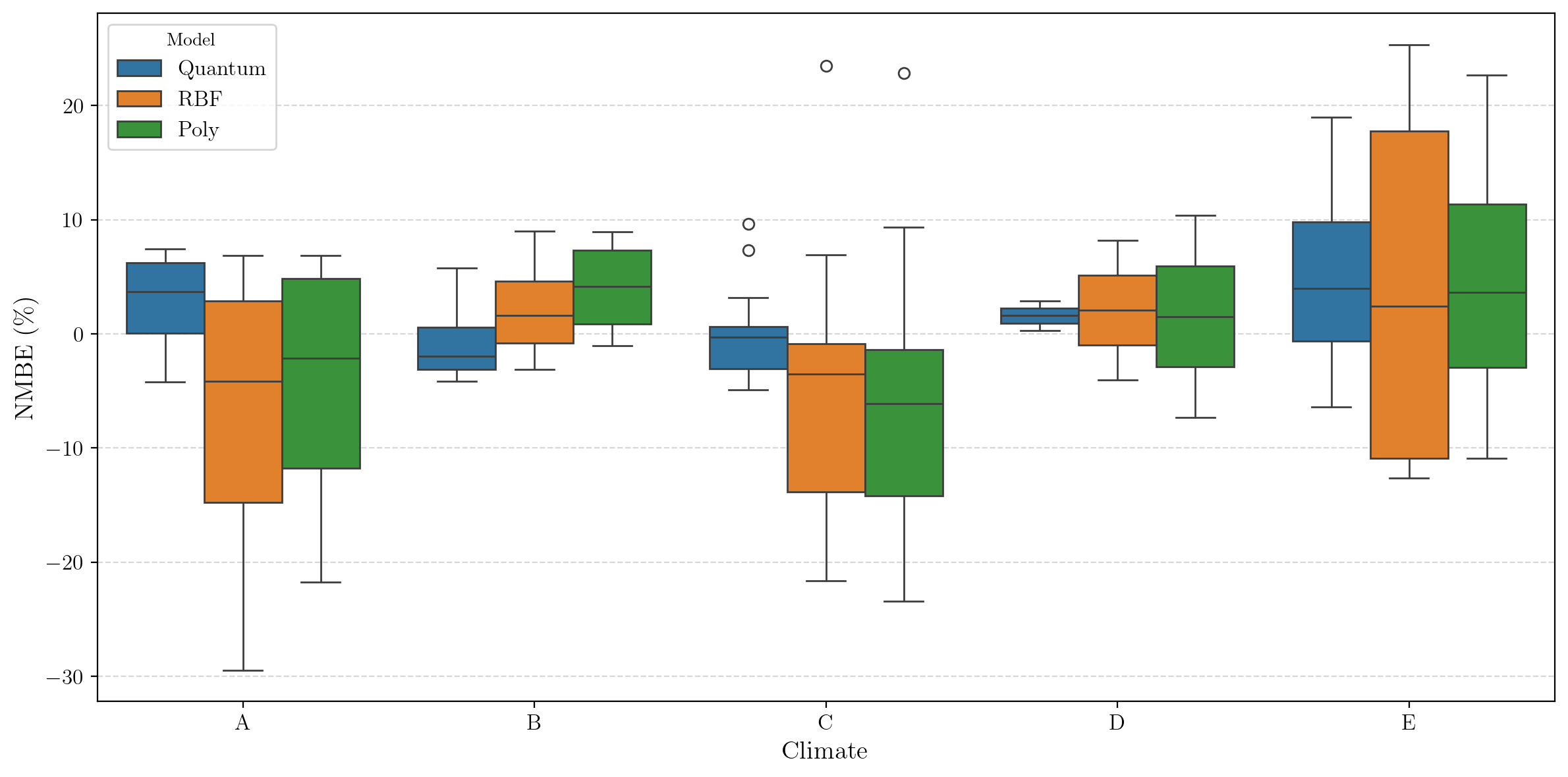}
    \caption{Normalized mean bias error (NMBE) for QK, RBF, and Poly kernels grouped by Köppen climate class.}
    \label{fig:3}
\end{figure}

Figure~\ref{fig:3} illustrates that QK also delivers the most balanced forecasts in terms of normalized mean bias error (NMBE). In tropical climates, the QK median hovers near \(+3\,\%\) with an IQR confined to \(\pm5\,\%\), whereas RBF and Poly exhibit pronounced underestimation, with medians near \(-5\,\%\) and \(-2\,\%\) and long lower tails to \(-30\,\%\) and \(-20\,\%\), respectively. Arid conditions shift QK to an almost neutral median (\(-1\,\%\)), contrasting with positive biases of about \(+3\,\%\) RBF and \(+5\,\%\) Poly.  Temperate sites again favor QK, centered at \(-1\,\%\) and within a \(\pm3\,\%\) IQR, while classical alternatives skew more negatively, with median of \(-4\,\%\) RBF and \(-6\,\%\) Poly. In snow regions, all models lean slightly positive, yet QK retains the narrowest spread (median \(+2\,\%\), IQR \(<2\,\%\)), whereas RBF and Poly approach \(+4\,\%\) with broader dispersion.  Polar stations pose the greatest challenge, but QK still limits its median to \(+4\,\%\) and IQR to \(\pm6\,\%\), while classical kernels range from \(-12\,\%\) to \(+18\,\%\) (RBF) and \(-10\,\%\) to \(+22\,\%\) (Poly). These biased statistics corroborate the earlier error and goodness‐of‐fit analyses, indicating that QK not only lowers overall error magnitude but also mitigates systematic deviation across diverse climatic contexts.

\begin{figure}[!htb]
    \centering
    \includegraphics[width=0.8\linewidth]{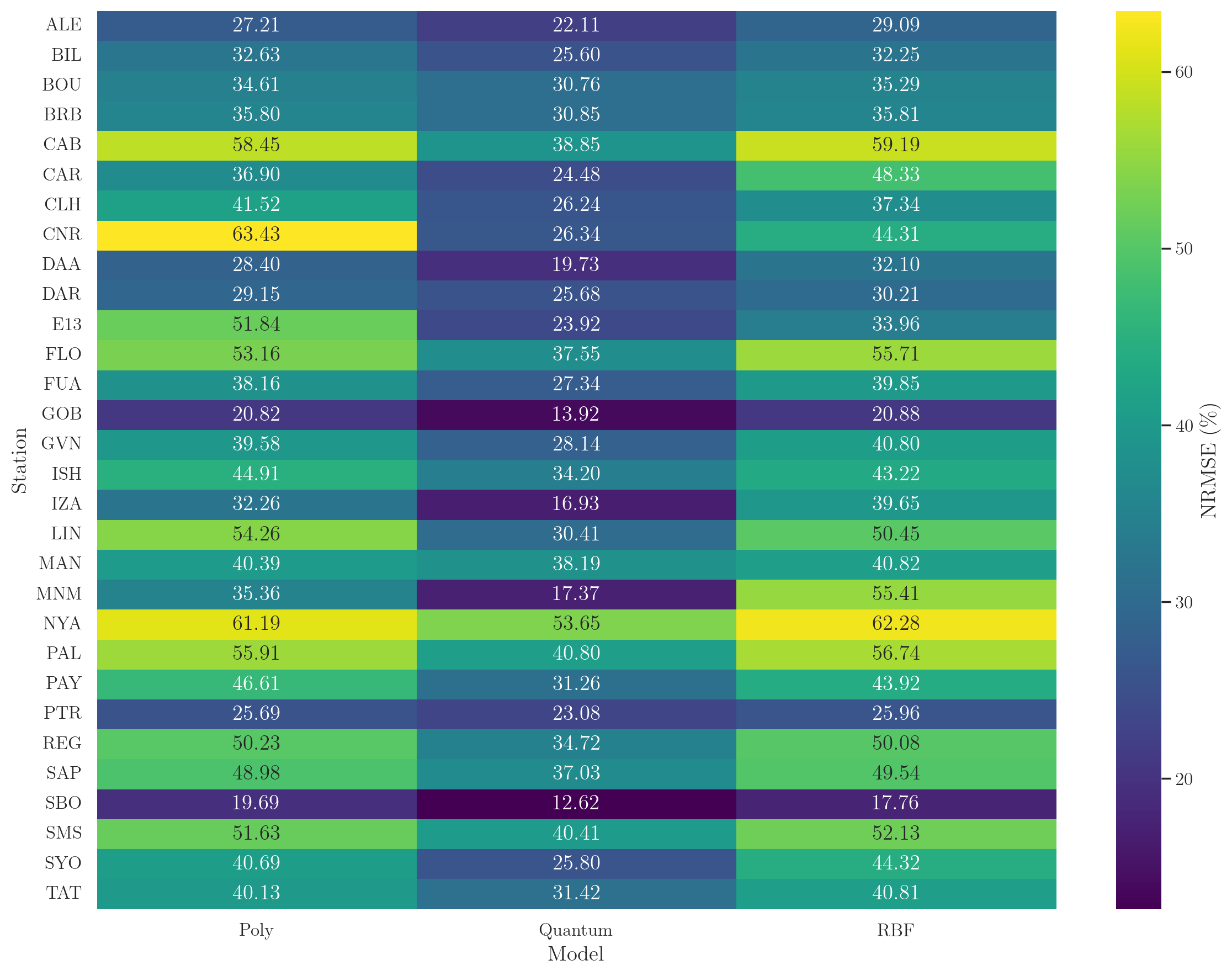}
    \caption{Station-level NRMSE for QK, RBF, and Poly kernels.}
    \label{fig:4}
\end{figure}

The station-level heatmap in Figure~\ref{fig:4} confirms QK’s systematic advantage.  Among the 30 stations with complete data, QK yields the lowest NRMSE throughout, producing median reductions of 29\,\% relative to Poly and 31\,\% relative to RBF. The station with the lowest NRMSE was SBO, where the QK achieved \(12.6\,\%\), compared with \(19.6\,\%\) for Poly and \(17.8\,\%\) for RBF. In contrast, the highest errors occurred at NYA, with \(53.65\,\%\) for QK, \(61.19\,\%\) for Poly, and \(62.28\,\%\) for RBF. These results highlight that, even in the most challenging locations, the quantum kernel consistently attains lower prediction errors than the classical counterparts. Figure~\ref{fig:4} clearly shows that the QK model surpasses both classical kernel approaches across all stations, demonstrating superior generalization and robustness to regional variability.

\begin{figure}[!htb]
    \centering
    \includegraphics[width=0.8\linewidth]{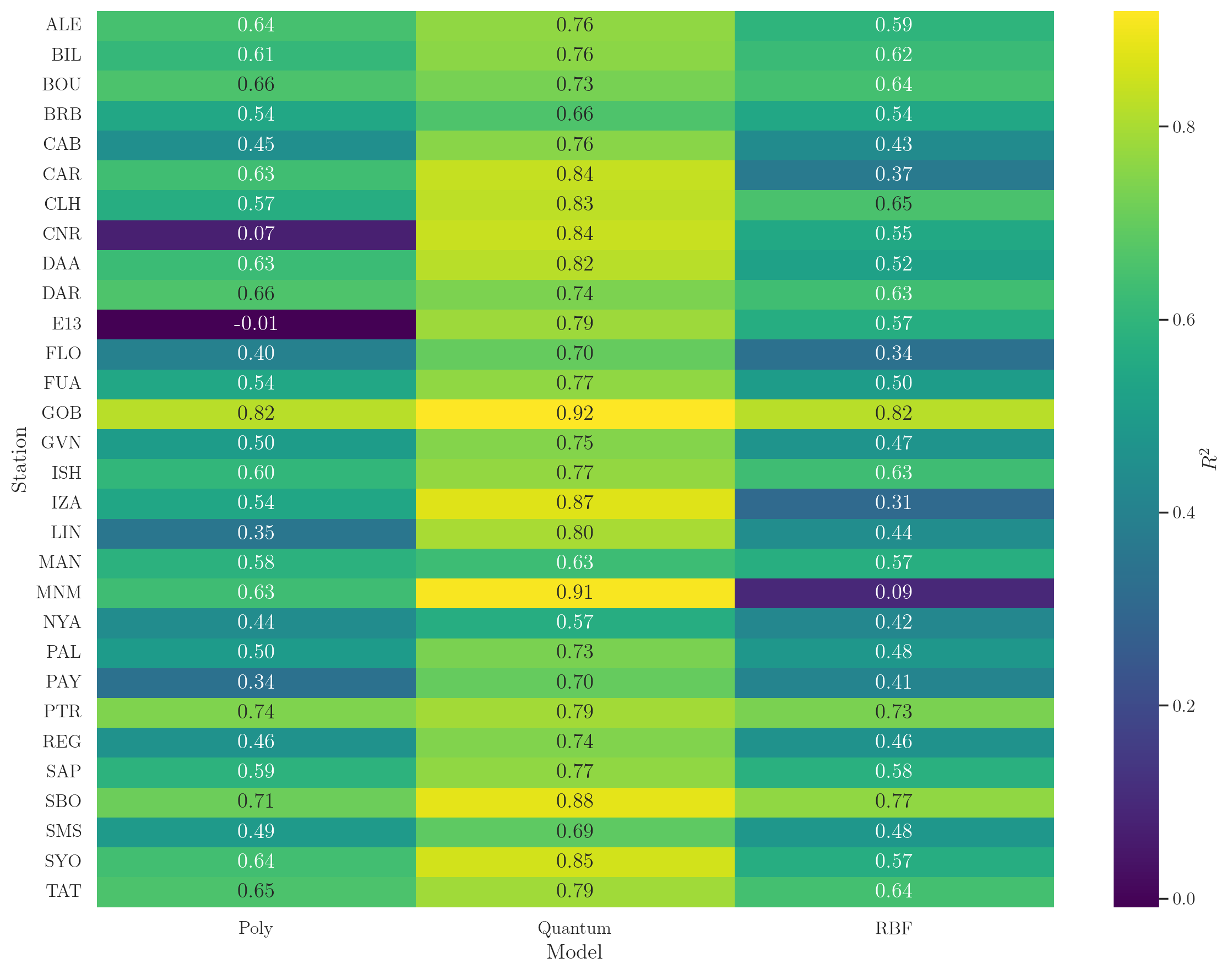}
    \caption{Station-level \(R^{2}\) for QK, RBF, and Poly kernels.}
    \label{fig:5}
\end{figure}

The heatmap of \(R^{2}\) in Figure~\ref{fig:5} shows that the QK delivers the greatest explanatory power at every station, outperforming Poly and RBF baselines without exception. Median \(R^{2}\) rises from roughly 0.57 for the classical kernels to 0.76 with QK. Large site-specific gains appear at LIN (0.80 versus 0.35 and 0.44), IZA (0.87 against 0.54 and 0.31), and MNM (0.91 compared with 0.63 and 0.09). The lowest \(R^{2}\) for QK was found in NYA, with an \(R^{2}\) of 0.57, while the lowest for Poly was E13 with -0.01 and for RBF was MNM with 0.09.  

\begin{figure}[!htb]
    \centering
    \includegraphics[width=0.8\linewidth]{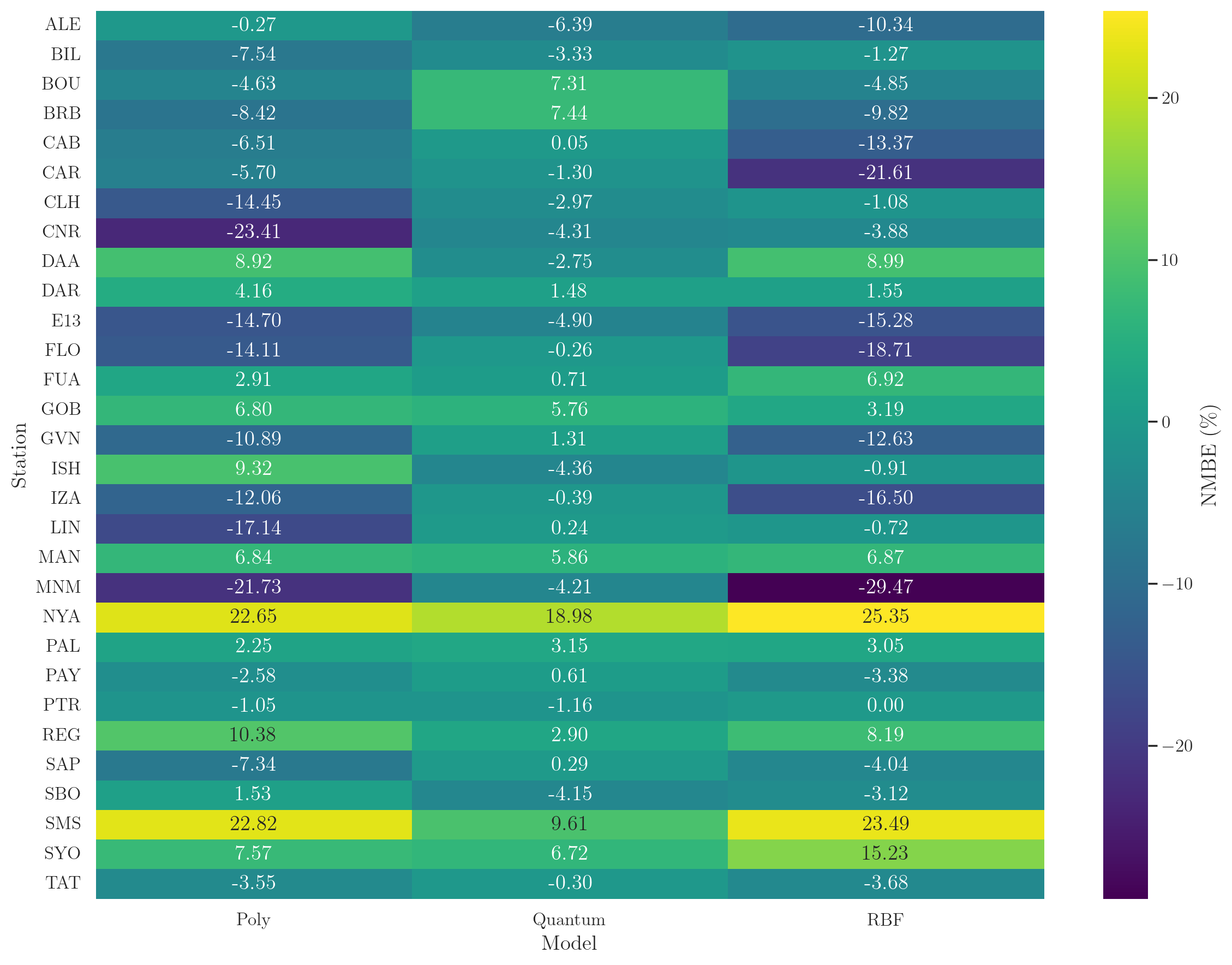}
    \caption{Station-level NMBE for QK, RBF, and Poly kernels.}
    \label{fig:6}
\end{figure}

Finally, Figure~\ref{fig:6} highlights station-level bias patterns.  Across 30 stations, QK’s median NMBE is essentially neutral, with a median of \(\approx -0.3\,\%\). In contrast, Poly and RBF exhibit systematic negative drift—medians of \(-3.1\,\%\) and \(-1.2\,\%\). Their drift extends from severe underestimation at CNR (\(-23.4\,\%\), Poly) and MNM (\(-29.5\,\%\), RBF) to large overestimation at NYA and SMS (\(+22\text{ and }25\,\%\), respectively). Even at sites where all models show positive bias (e.g., DAA, GOB, FUA), QK remains closest to zero. There are only three stations where the NMBE for QK is more biased than the classical kernels. These stations are BOU with an NMBE of 7.31\% for QK, PAL with 3.15\% QK, and SBO with -4.15\% QK.

\section{Discussion}


Quantum kernel computations presented in this study used PennyLane’s \texttt{lightning.\allowbreak qubit} simulator, representing an ideal, noise-free quantum environment. Implementing this approach on today's quantum hardware could be challenging due to significant noise, decoherence, and measurement errors inherent to Noisy Intermediate-Scale Quantum (NISQ) devices. Furthermore, the quantum circuit employed here, particularly due to the QFT, has a non-negligible depth, potentially compounding these practical difficulties. Nevertheless, the promising performance demonstrated in this study highlights the potential of quantum kernels, motivating future research efforts and technological developments toward practical quantum machine learning applications.

Both classical and quantum kernels were trained and evaluated on identical datasets (approximately 1900 training windows). Given the relatively limited training set size, classical kernels exhibited poorer performance, possibly indicating sensitivity to training set size or difficulty capturing the data's underlying periodicity. Conversely, the quantum kernel consistently achieved higher accuracy under identical conditions. This observation suggests that the structure of the QFT-based quantum kernel inherently captures the periodic characteristics of solar irradiance data more effectively, or that its design is particularly advantageous when dealing with smaller datasets. Further investigation involving varying dataset sizes could clarify the observed differences.
 
The favorable results obtained with the quantum kernel likely stem from the QFT's inherent ability to represent periodic or cyclic data efficiently within the quantum feature space. Given the pronounced periodic patterns characteristic of solar irradiance data, the quantum embedding employed here is especially suited to exploit these frequency-domain structures effectively. This suitability underscores the relevance and strength of the QFT-based embedding for similar time-series forecasting problems. The strong predictive accuracy demonstrated by the quantum kernel could partially come from the protective layer's ability to generate more expressive and distinctive quantum embeddings compared to purely amplitude-based encoding. This particular design choice appears especially well-suited to capturing the inherent periodicities of solar irradiance data.

Across all three performance metrics, the QK consistently outperforms both classical references: the RBF and Poly kernels. Climate‐class analyses reveal that QK reduces median NRMSE by 7–15 percentage points relative to the classical models, elevates median \(R^{2}\) by 0.10–0.30, and keeps NMBE within \(\pm5\,\%\) even in challenging polar and snow regimes.  Moreover, QK’s interquartile ranges are uniformly narrower, indicating greater stability and less susceptibility to extreme errors or biases across diverse atmospheric conditions. Station‐level heatmaps reinforce these findings: QK delivers the lowest NRMSE and the highest \(R^{2}\) at every site with complete data, while preserving an essentially unbiased NMBE distribution with the tightest interquartile spread. Classical kernels, by contrast, frequently show substantial underestimation or overestimation, with NMBE drifts of \(\pm20\,\%\) or more—and exhibit pronounced error dispersion. Whereas the classical models show broader tails and systematic under/overestimation at several stations, the quantum kernel suppresses these extremes, indicating a more stable, frequency-aware representation across heterogeneous geographies. Still, a handful of locations subject to rapid transients retain elevated maxima and wider spreads, suggesting headroom for robustness enhancements under highly variable cloud conditions.

\begin{figure}[!htb]
    \centering
    \includegraphics[width=0.9\linewidth]{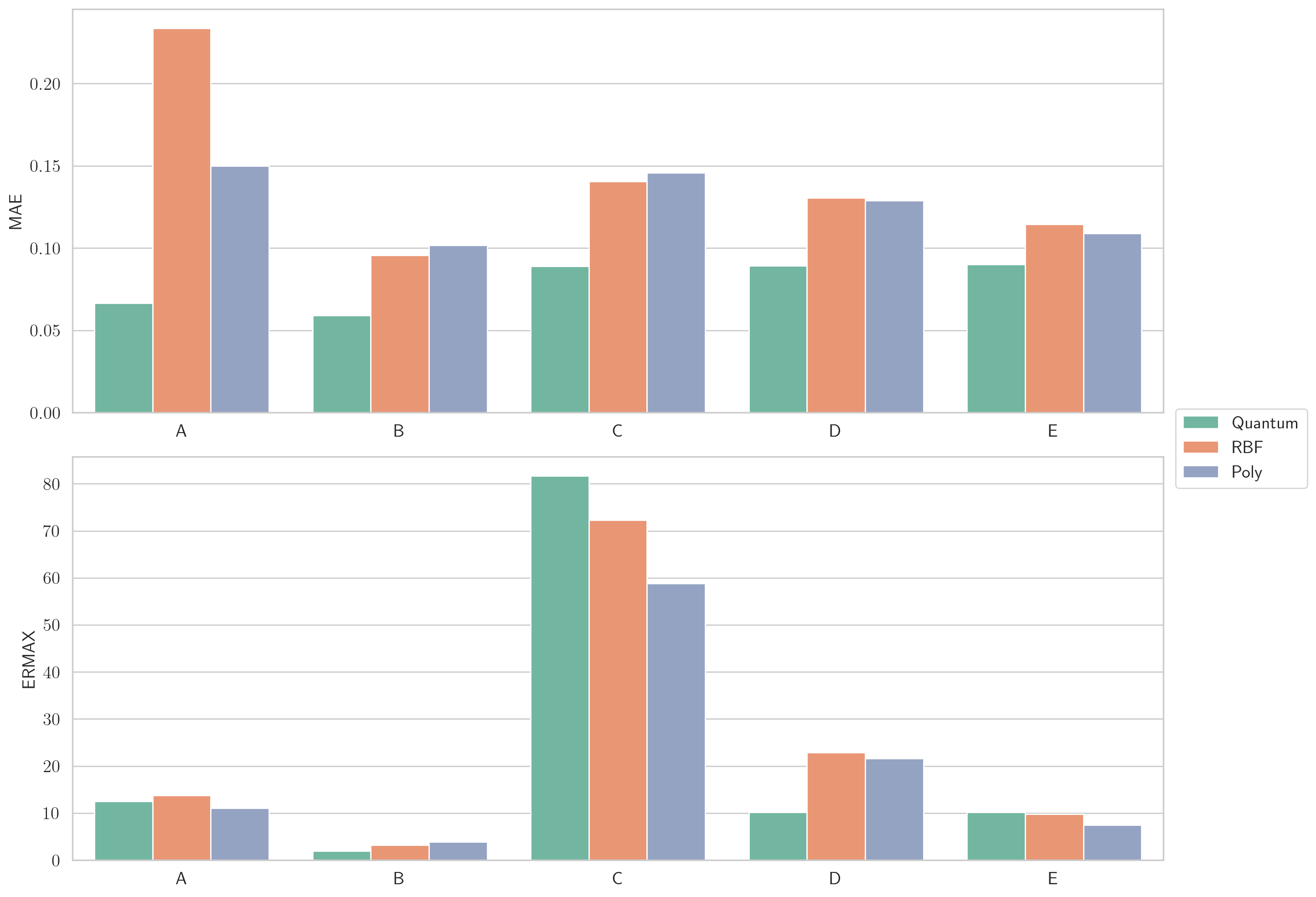}
    \caption{MAE and ERMAX for the three models evaluated.}
    \label{fig:disc}
\end{figure}

Figure~\ref{fig:disc} complements the main results by examining average accuracy, represented by the Mean Absolute Error (MAE), and worst-case behaviour, captured by the Maximum Error Ratio (ERMAX), across Köppen classes. The MAE quantifies the mean magnitude of absolute deviations between predictions and observations, providing a measure of overall forecasting precision. In contrast, ERMAX identifies the largest relative deviation observed, offering insight into the model's stability and resilience under extreme conditions. Together, these metrics allow a more complete evaluation of both the central tendency and the extremes of model performance across different climatic regimes. In the MAE panel, the QK attains the lowest error in every climate, with particularly pronounced gains in tropical and temperate–polar (C–E) regimes. Relative to RBF and Poly baselines, reductions in MAE typically fall in the 30–60\% range—substantial in class A and still material in the drier and more stable class B. These patterns indicate that the frequency-aware embedding is capturing diurnal/seasonal structure and short-term variability more faithfully than classical kernels, translating into uniformly tighter central error across heterogeneous meteorological regimes.

The ERMAX panel in Figure~\ref{fig:disc}, however, makes clear that the model is not yet uniformly robust to extremes. While QK matches or improves the maxima in several climates (e.g., low outliers in B and D), temperate sites (C) exhibit large peak errors for all methods, and the quantum kernel does not suppress those spikes; this points to vulnerability during abrupt transients such as cloud-edge passages or rapidly evolving aerosol/precipitation events. Several avenues could mitigate this: (i) incorporating exogenous, fast-changing cues (sky images, satellite cloud motion vectors, aerosol optical depth) into the feature-weighted kernel mix; (ii) multi-scale windows or wavelet/QFT hybrids to capture sharp ramps; (iii) training with robust or tail-sensitive objectives (Huber, quantile, or CVaR-style losses) to explicitly penalize extremes; and (iv) regime-aware weighting that adapts kernel contributions by time-of-day and cloudiness. Therefore, the quantum kernel clearly improves average performance, but controlling rare, large deviations remains an open and tractable target for the next iteration.

\section{Conclusions}


This study introduced a Quantum Fourier Transform (QFT)–based quantum kernel for short-term solar irradiance forecasting, combining amplitude encoding, a QFT stage, and a rotational protective layer within a kernel-ridge framework and fusing per-feature kernels via convex weights. Across all five Köppen climate classes and at the station level, the proposed kernel consistently outperformed radial-basis and polynomial references, yielding lower \(n\mathrm{RMSE}\), higher \(R^{2}\), and near-zero \(n\mathrm{MBE}\). Complementary analyses of MAE and worst-case error (ERMAX) further showed tighter average accuracy and a reduction—though not elimination—of extreme events. These findings indicate that embedding windowed irradiance histories in a frequency-aware quantum feature space effectively captures diurnal/seasonal structure and short-term variability. 

Beyond solar irradiance, the methodology is generic: the same kernel construction and feature-weighted fusion can be applied to other periodic or quasi-periodic time-series tasks (e.g., wind nowcasting, building/electricity load, air-quality and wave/tide prediction, traffic flows), extended to multi-step horizons (direct or recursive strategies), adapted to spatio-temporal settings by incorporating neighboring stations or satellite pixels as additional kernels, and repurposed for anomaly detection or regime classification via one-class/large-margin kernel methods. It also lends itself to probabilistic forecasting by pairing kernel-ridge outputs with quantile objectives or conformal prediction on residuals, and to data-rich settings by integrating exogenous cues (sky imagers, cloud-motion vectors, clear-sky models) as separate kernels in the mixture. 

Some limitations should be acknowledged. Classical RBF/Polynomial hyperparameters $(\gamma, r, d)$ are fixed;
only the convex feature–mixing weights and the KRR ridge parameter $\alpha$ are tuned on the
validation set (same budget for all models). Consequently, the classical results are
reference baselines, not fully tuned classical state-of-the-art. All quantum‐kernel results are obtained on a noiseless state-vector simulator with 5-qubit circuits corresponding to 32-sample windows, so device noise, readout errors, and state-preparation overheads are not reflected in the reported accuracy. The QFT pathway assumes equally spaced samples and power-of-two window lengths due to amplitude encoding with L2 normalization, which may limit portability to other cadences or horizons without resampling or padding. A protective layer is required to avoid trivial cancellation of QFT/QFT$^{\dagger}$, constraining circuit design choices that are not exhaustively explored here. Model selection relies on a single chronological train/validation split with a modest optimization budget; stricter validation-only procedures (e.g., blocked or rolling cross-validation) would further reduce selection bias and yield tighter uncertainty estimates. Classical reference baselines (RBF and polynomial kernels) are used in fixed, untuned configurations; consequently, gaps reported against these models should be interpreted as comparisons to reference baselines rather than state-of-the-art alternatives.

Looking forward, several extensions appear promising. A parameterized QFT kernel can be realized without altering the QFT itself by inserting a small, trainable phase layer between the QFT and the protective layer; alternatively, parameterization can be confined to the protective layer via a compact set of trainable rotations. Both variants could allow for better alignment with the dataset. The protective layer can also be diversified through entangling variational patterns, data reuploading schemes, etc. Moreover, execution on NISQ hardware, with lightweight error-mitigation baselines, would quantify resilience beyond ideal simulation and inform circuit simplifications compatible with device depth and noise.

\section*{Acknowledgments}

Mr. Rodríguez would like to acknowledge the PhD scholarship from ANID - Subdirección de Capital Humano/Doctorado Nacional 2023-21230987. 

\bibliographystyle{model3-num-names}
\bibliography{References.bib}

\clearpage
\appendix
\section{Derivation of the quantum kernel expression}\label{appA}

This appendix provides a detailed mathematical derivation of the quantum kernel expression employed throughout the paper. The quantum kernel, expressed as the inner product between quantum-embedded data points, is explicitly derived for arbitrary time windows $W_s^{(i)}$ and $W_s^{(j)}$. This derivation highlights how amplitude encoding and Quantum Fourier Transform (QFT) operations produce the kernel formula presented in the main text.
\\
Let $W_s^{(i)}$ and $W_s^{(j)}$ with $i = 1,2,...,N_1$ and $j = 1,2,...,N_2$ be two arbitrary windows (train and/or test windows), the goal is to simplify the expression of the quantum kernel value between these two windows.
\\
\\
$W_s^{(i)}$ is the i-th window corresponding to feature s. Throughout the derivation, the most general case is adopted by using n qubits and setting $N = 2^n$.
\\
\begin{align}
\begin{split}   
    k(W_s^{(i)}, W_s^{(j)}) &= |\bra{0}U^\dagger(W_s^{(i)})U(W_s^{(j)})\ket{0} |^2  \\
    &=  |\bra{0}A^\dagger(W_s^{(i)})QFT^\dagger V^\dagger(W_s^{(i)})V(W_s^{(j)})QFT A(W_s^{(j)})\ket{0} |^2 \\
    &=  |\sum^{N-1}_{p=0}\Tilde{y}_p^{(i,s)}\bra{p}V^\dagger(W_s^{(i)})V(W_s^{(j)})\sum^{N-1}_{k=0}y_k^{(j,s)}\ket{k}|^2 \\
    &= |\sum_{p, k}\Tilde{y}_p^{(i,s)}y_k^{(j,s)}\bra{p}V^\dagger(W_s^{(i)})V(W_s^{(j)})\ket{k}|^2
\end{split}
\end{align}
\\
Where $y^{(j,s)}_k = \frac{1}{\sqrt{N}} \sum_{v=0}^{N-1}(W_s^{(j)})_v e^{2\pi i \frac{vk}{N}}$ and $\Tilde{y}_p^{(i,s)} = \frac{1}{\sqrt{N}} \sum_{v=0}^{N-1}(W_s^{(i)})_v e^{-2\pi i \frac{vp}{N}}$
\\
\\
It follows that $\bra{p}V^\dagger(W_s^{(i)})V(W_s^{(j)})\ket{k}$ is the matrix element
\begin{equation}
    (V^\dagger(W_s^{(i)})V(W_s^{(j)}))_{pk} = R_{pk}^{(i,j,s)}
\end{equation}
Therefore, 

\begin{align}
\begin{split}  
    k(W_s^{(i)}, W_s^{(j)}) &= |\sum_{k,p} \sigma_{pk}^{(i,j,s)}R_{pk}^{(i,j,s)}|^2 \\
    &= |Tr(\sigma^{(i,j)}_s R^{(i,j)}_s)|^2
\end{split}
\end{align}

Where $\sigma^{(i,j)}_s$ is the matrix of element $\sigma_{pk}^{(i,j,s)} = \Tilde{y}_p^{(i,s)}y_k^{(j,s)}$ and $R^{(i,j)}_s = V^\dagger(W_s^{(i)})V(W_s^{(j)})$

\clearpage

\section{Continuation of the kernel value derivation with the QFT}\label{appB}

This appendix continues the derivation of the kernel value expression by detailing the mathematical action of the Quantum Fourier Transform (QFT) that is built into the quantum circuit. 
\\
Consider the vector\begin{align}
    y^{(j,s)} &= \begin{bmatrix}
           y^{(j,s)}_0 \\
           y^{(j,s)}_1 \\
           \vdots \\
           y^{(j,s)}_{N-1}
         \end{bmatrix}
  \end{align}
\\
\\ 
By definition of the Fourier coefficients :
\begin{align}
    y^{(j,s)} &= \frac{1}{\sqrt{N}} \begin{bmatrix}
           \sum_{l=0}^{N-1} (W_s^{(j)})_l e^{2 \pi i \frac{l\times 0}{N}}\\
           \sum_{l=0}^{N-1} (W_s^{(j)})_l e^{2 \pi i \frac{l \times 1}{N}} \\
           \vdots \\
           \sum_{l=0}^{N-1} (W_s^{(j)})_l e^{2 \pi i \frac{l (N-1)}{N}}
         \end{bmatrix}
         &= \frac{1}{\sqrt{N}} \sum_{l=0}^{N-1} (W_s^{(j)})_l \begin{bmatrix}
            \omega_N^{l \times 0 }\\
            \omega_N^{l \times 1} \\
           \vdots \\
            \omega_N^{l(N-1)}
         \end{bmatrix}
  \end{align}
\\
\\ 
where $\omega_N = e^{\frac{2 \pi i }{N}}$.
\\
\\
Consequently, 
\begin{align}
\begin{split}
    \sigma^{(i,j)}_s &= \Tilde{y}^{(i,s)}y^{(j,s)^T}  \\
    &=  \frac{1}{N} \sum_{l, v} (W_s^{(i)})_l (W_s^{(j)})_v\begin{bmatrix}
            \omega_N^{-l \times 0 }\\
            \omega_N^{-l \times 1} \\
           \vdots \\
            \omega_N^{-l(N-1)}
         \end{bmatrix}\begin{bmatrix}
            \omega_N^{v \times 0 }, 
            \omega_N^{v \times 1} 
           \hdots 
            ,\omega_N^{v(N-1)}
         \end{bmatrix} \\
    &=\frac{1}{N} \sum_{l, v} \beta_{lv}^{(i,j,s)} \begin{bmatrix}
    1 & \omega_N^{v} & \dots  & \omega_N^{v(N-1)} \\
    \omega_N^{-l} & \omega_N^{v-l} &   &  \\
    \vdots &   & \ddots & \vdots \\
    \omega_N^{-l(N-1)} &  & \dots  & \omega_N^{(v-l)(N-1)}
\end{bmatrix} \\
     &= \frac{1}{N} \sum_{l, v} \beta_{lv}^{(i,j,s)} \Omega^{(l,v)} 
\end{split}
\end{align}

Where $\beta_{lv}^{(i,j,s)} =(W_s^{(i)})_l (W_s^{(j)})_v $ and $\Omega^{(l,v)} = \begin{bmatrix}
    1 & \omega_N^{v} & \dots  & \omega_N^{v(N-1)} \\
    \omega_N^{-l} & \omega_N^{v-l} &   &  \\
    \vdots &   & \ddots & \vdots \\
    \omega_N^{-l(N-1)} &  & \dots  & \omega_N^{(v-l)(N-1)}
\end{bmatrix}$
\\
\\ 
\\
Therefore,
\begin{equation}
    k(W_s^{(i)}, W_s^{(j)}) = |\frac{1}{N} \sum_{l, v} \beta_{lv}^{(i,j,s)} Tr(\Omega^{(l,v)}R^{(i,j)}_s)|^2
\end{equation}

\clearpage

\section{Construction of the Rotational Protective Layer \(V(x)\)}\label{appC}

This appendix completes the kernel derivation by computing \(R^{(i,j)}_{s}=V^{\dagger}\!\bigl(W^{(i)}_{s}\bigr)V\!\bigl(W^{(j)}_{s}\bigr)\), which describes the action of the rotational protective layer (used in this study) that follows the QFT in the circuit presented in Figure \ref{fig:QFTkernel}.
\newline
\newline
\underline{\textbf{Using the rotational encoding variation for $V(\cdot)$:}}
\newline
\newline
\indent If n qubits are used, one window must have size $2^n =N$.

Performing an Euclidean division, $2^n = an + b$; after some calculation, the following quantity can be computed:

\begin{align}
\begin{split}
    R^{(i,j)}_s &= V^\dagger(W_s^{(i)})V(W_s^{(j)})  \\
    &= \Bigg[  \bigotimes_{m=1}^{n-1} \Bigg[  \prod_{\text{r odd, r' odd}}^{a} R_X^\dagger\left( (\mathcal{A}^{(i)}_m)_r)\right) R_X\left( (\mathcal{A}^{(j)}_m)_{r'})\right)  \\ 
    & + \prod_{\text{r odd, r' even}}^{a}R_X^\dagger\left( (\mathcal{A}^{(i)}_m)_r)\right) R_Y\left( (\mathcal{A}^{(j)}_m)_{r'})\right) \\
    & + \prod_{\text{r even, r' odd}}^{a}R_Y^\dagger\left( (\mathcal{A}^{(i)}_m)_r)\right) R_X\left( (\mathcal{A}^{(j)}_m)_{r'})\right) \\
    & + \prod_{\text{r even, r' even}}^{a}R_Y^\dagger\left( (\mathcal{A}^{(i)}_m)_r)\right) R_Y\left( (\mathcal{A}^{(j)}_m)_{r'})\right) \Bigg] \Bigg] \\
    & \otimes \Bigg[ \prod_{\text{r odd, r' odd}}^{a+b} R_X^\dagger\left( (\mathcal{A}^{(i)}_m)_r)\right) R_X\left( (\mathcal{A}^{(j)}_m)_{r'})\right)  \\
    & + \prod_{\text{r odd, r' even}}^{a+b}R_X^\dagger\left( (\mathcal{A}^{(i)}_m)_r)\right) R_Y\left( (\mathcal{A}^{(j)}_m)_{r'})\right) \\
    & + \prod_{\text{r even, r' odd}}^{a+b}R_Y^\dagger\left( (\mathcal{A}^{(i)}_m)_r)\right) R_X\left( (\mathcal{A}^{(j)}_m)_{r'})\right) \\
    & + \prod_{\text{r even, r' even}}^{a+b}R_Y^\dagger\left( (\mathcal{A}^{(i)}_m)_r)\right) R_Y\left( (\mathcal{A}^{(j)}_m)_{r'})\right) \Bigg]  \\
\end{split}
\end{align}
with r,r' $\geq 1$.

Where $\mathcal{A}_i = \{ \{ (W_s^{(i)})_v \mid v=1,2,...,a \}, \{ W_s^{(i)})_v \mid v=(a+1),...,2a \},..., \{W_s^{(i)})_v \mid v=((n-2)a +1),...,(n-1)a  \},\{W_s^{(i)})_v \mid ((n-1)a +1),...,(na +b) \} \} \xrightarrow{ }$ this order is fixed.

\end{document}